\documentclass[preprint,12pt,authoryear]{elsarticle}
\journal{Information Sciences}

\usepackage{rotating}
\usepackage{lipsum}
\usepackage{mwe}

\usepackage{latexsym}
\usepackage{amsmath}
\usepackage{graphicx}
\usepackage{multirow}
\usepackage{algorithm}
\usepackage[noend]{algpseudocode}
\usepackage{wrapfig}
\usepackage{changepage}

\usepackage{setspace} 
\usepackage{subfig} 

\newlength\myindent
\usepackage{orcidlink}

\usepackage{array}
\newcolumntype{L}{>{\arraybackslash}m{7cm}}
\newcolumntype{n}{>{\arraybackslash}m{1.9cm}}

\usepackage{booktabs}

\begin{document}
\begin{frontmatter}
\title{Contrastive Federated Learning with Tabular Data Silos}

\author [a,d]{Achmad Ginanjar}
\author[a]{Xue Li}
\author[a]{Priyanka Singh}
\author[b]{Wen Hua}
\author[c]{Jiaming Pei}

\affiliation[a]{organization={School of Electrical Engineering and Computer Science},
            addressline={The University of Queensland},
            city={Brisbane},
            postcode={4067},
            state={Queensland},
            country={Australia}}

\affiliation[b]{organization={Department of Computing,Hong Kong Polytechnic University},
            addressline={Hong Kong Polytechnic University},
            country={Hong Kong}}
\affiliation[c]{organization={The University of Sydney},
            city={Sydney},
            country={Australia}}
\affiliation[d]{organization={Indonesia Tax Office},
            city={Jakarta},
            country={Indonesia}}
\begin{abstract}

Learning from vertical partitioned data silos is challenging due to the segmented nature of data, sample misalignment, and strict privacy concerns. Federated learning has been proposed as a solution. However, sample misalignment across silos often hinders optimal model performance and suggests data sharing within the model, which breaks privacy.  Our proposed solution is Contrastive Federated Learning with Tabular Data Silos (CFL), which offers a solution for data silos with sample misalignment without the need for sharing original or representative data to maintain privacy. CFL begins with local acquisition of contrastive representations of the data within each silo and aggregates knowledge from other silos through the federated learning algorithm.
Our experiments demonstrate that CFL solves the limitations of existing algorithms for data silos and outperforms existing tabular contrastive learning. CFL provides performance improvements without loosening privacy. The code can be accessed online \cite{cflCode}.
\end{abstract}



\end{frontmatter}


\section{Introduction}
The existence of data silos across organizations presents significant challenges for collaborative learning. These challenges come from the segmented nature of data (partition), sample misalignment ( presence of non-identically and independently distributed data/non IID, label costlines), and strict privacy concerns. Various factors contribute to these problems, including differences in data collection methods, time-related dependencies, spatial dependencies, and law. 

Research within this area partially solves the above problems. For instance, current collaborative learning models for vertically partitioned data/data silo \cite{FEDCVT} often require data sharing. This approach is not feasible due to privacy concerns. To address these challenges, a new approach is needed.

In this paper, we focus on collaborative learning, which addresses the following key challenges:
\begin{itemize}
    \item \textbf{Vertically partitioned tabular data silos}: Each silo possesses specific data columns based on its functions. However, they  are connected via quasi-identifiers \cite{Motwani2007}. Collaborative learning is challenging because the data is fragmented. 
    \item  \textbf{Sample misalignment}: The data collected by each silo may be imbalanced due to non IID, label costlines. Both lead to sample misalignment when performing collaborative learning on vertically partitioned data due to join operations, which led to a low-performance model.
    \item \textbf{Privacy constraints}: Our research enforces strict privacy constraints, preventing the transfer of original data and disallowing the use of third-party agents. These requirements are essential in industries such as the government \cite{Zhou2020}.
\end{itemize}

Federated Learning (FL) \cite{McMahan2017,acmFLPrivacy, FlVerticalSemiSupervised} and Contrastive Learning (CL) \cite{Gutmann2010, acmDataAugmented,representationSurvey} have emerged as potential solutions. FL is a multi-agent collaborative learning algorithm that enables model training across multiple silos without compromising privacy.  Although it covers vertical partition, vertical federated learning (VFL) struggles to retain strict privacy constraints. CL is an algorithm to self-enhance data by encoding existing data for self-optimization and improved supervised learning performance, especially for small sample sizes. In CL, the loss is calculated based on two different sets of augmented data from unknown features. CL is an image-based model and is not designed to handle collaborative learning. To our knowledge, current FL and CL approaches, such as FedCVT \cite{FEDCVT} and MOON \cite{Li2021contrastiveFL}, do not effectively handle the challenges posed by vertical tabular data silos with sample misalignment without partially sharing original data.
\begin{figure}[]
    \centering
    \includegraphics[width=1\linewidth]{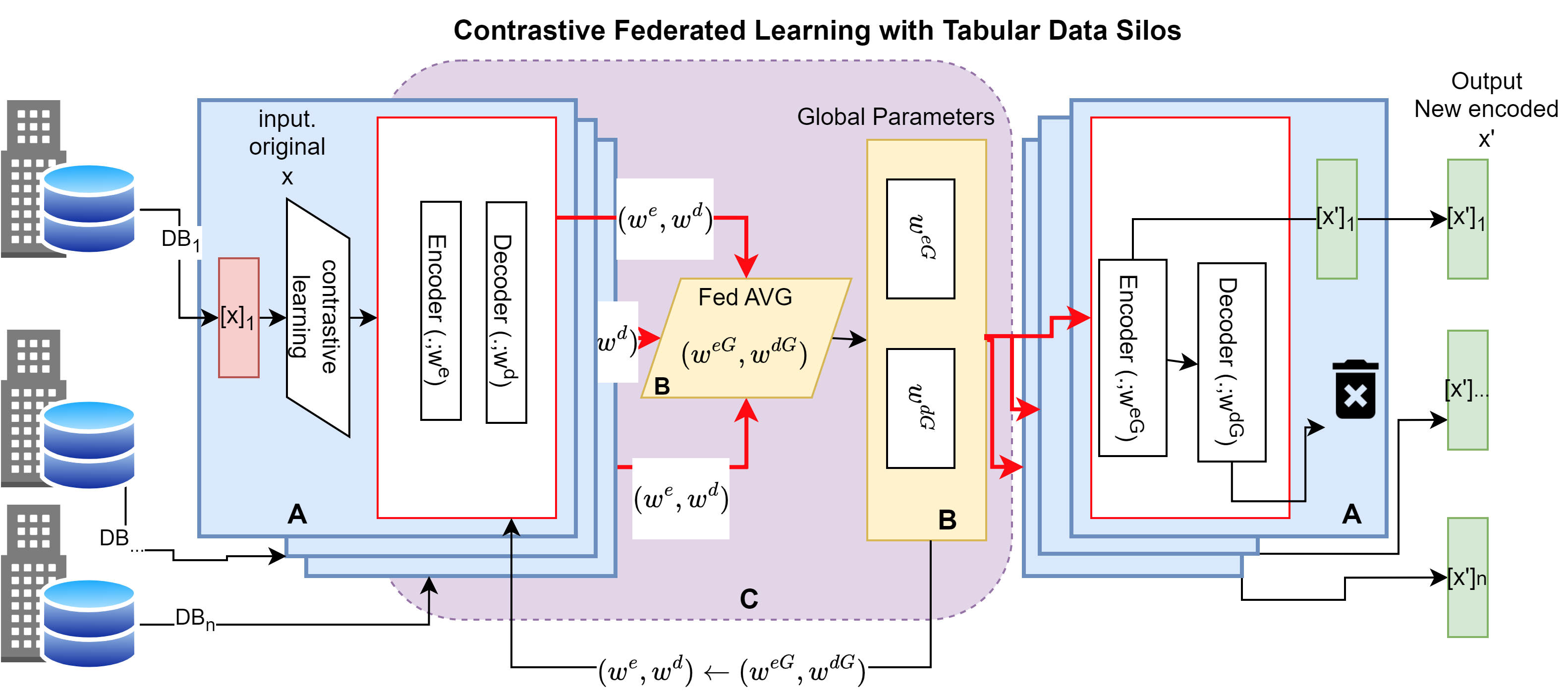}
    \caption{Contrastive Federated Learning with Tabular Data Silos. The (\textbf{A}) areas are local contrastive learning, the (\textbf{B}) area is server learning, and the (\textbf{C}) area is the objects involved in federated learning. $[x]$ is the original data matrix and $[x']$ is the output matrix for supervised inferences.  }
    \label{fig:CFL}
\end{figure}

\begin{figure}
    \centering
    \includegraphics[width=1\linewidth]{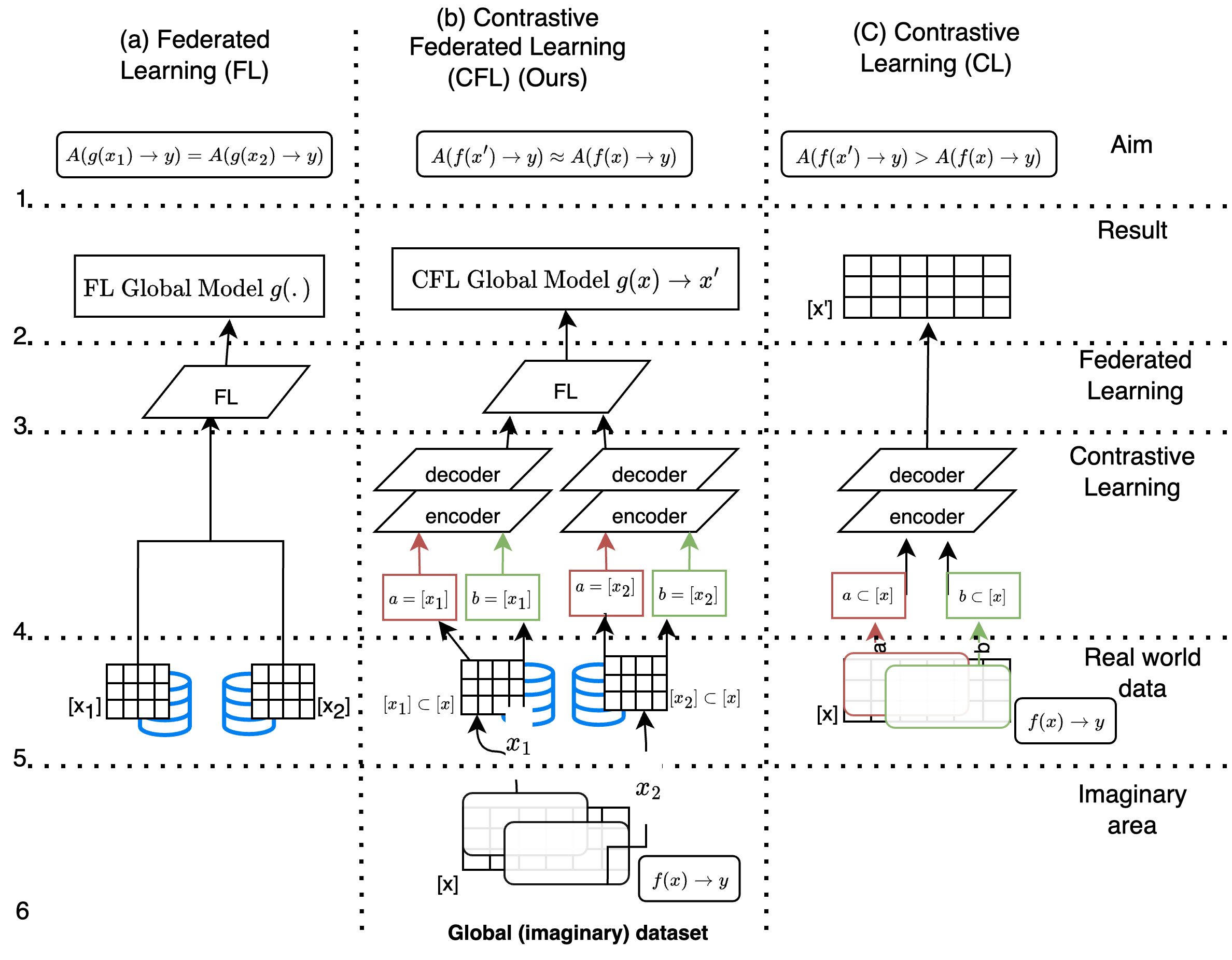}
    \caption{CFL leverages the power of contrastive learning (CL) to find similarities between two data slices and federated learning (FL) to share knowledge between silos. Part (b.6) shows where the CFL problem begins. (b.6) is similar to (c.5), while (a.5) is similar to (b.5). The data in (b.5) are a slice, similar to in (c.4). The representation in (b.4) is a full tuple representation because it came from (b.5), which is a slice of (b.6). Slices (c.4) and (b.5) have different column name/features. $A(.)$ is evaluation function, $g(.)$ is the global model function, $f(.)$ is the local model function}
    \label{abstract}
\end{figure}

We propose Contrastive Federated Learning with Tabular Data Silos (CFL), a novel approach that combines the strengths of both FL and CL for tabular data silos with strict privacy constraints, see Figure \ref{fig:CFL}. Our CFL is driven by conditions such as VFL and hidden feature learning from CL, see Figure \ref{abstract}.  Our approach begins with local contrastive learning in each silo, resulting in a model with an encoder and decoder. These local models are then aggregated globally on a central server, taking advantage of the diverse knowledge from all silos. This is done by exchanging the parameters of both the encoder and decoder. The final model produces a refined encoder for each silo that effectively encodes new data to make better prediction tasks. CFL apply full matrix representation to adapt with tabular data, a Pearson reordering to introduce contextual relation, zero imputation to handle sample misalignment, and modified loss to speed up training. During inference, CFL serves as a pre-trained model for the client, as suggested by Tzinis \textit{et al}. \cite{Tzinis2021}. 

FL and CL within our CFL framework enable black-box learning. In CFL, features across different silos are kept hidden. Our CFL framework calculates a loss based on pairs of data that come from different features while maintaining anonymity, which stands in contrast to Vertical Federated Learning (VFL)  \cite{FlVerticalSemiSupervised}. This capability is made possible through the use of contrastive learning, which evaluates losses from pairs of data derived from different views. CFL displays characteristics similar to standard or non-vertical FL. Additionally, CFL can be improved by incorporating standard FL security methods; however, a discussion of these methods is not included in this paper.

The key contributions of this article can be summarized as follows:
\begin{itemize}
    \item We propose CFL to handle vertical partitioning, sample misalignment, and strict privacy constraints simultaneously. 
    \item We implement enhanced CL in our CFL with zero imputation to handle sample misalignment, silo full matrix representation to adopt tabular data, Pearson reordering to boost performance, and modified loss to speed up the model.
    \item We carry out extensive experiments to replicate the conditions in the real world, including class imbalance, size imbalance, and a combination of both.
\end{itemize}

Our unique mix approach enables CFL to perform well in extreme settings and outperform baseline models.

\section{Related Work   }
This section provides a comprehensive overview of existing research relevant to our work on Contrastive Federated Learning with Tabular Data Silos (CFL). We explore three key areas: learning on vertical data silos, addressing sample misalignment, and techniques for learning with label costliness.

\subsection{Learning on Vertical Data Silos}
Federated Learning (FL) \cite{McMahan2017} has emerged as a promising approach for collaborative learning while preserving data privacy. Recent advancements in FL have focused on improving network efficiency  \cite{Liu2019, Chen2020VAfl, Ji2023}, enhancing privacy guarantees  \cite{Lu2020, MoFan2022}, and addressing data quality issues \cite{Yang2018,Chen2019,Ramaswamy2019,Wang2020}. However, these methods often struggle with vertically partitioned data, which is common in enterprise settings.

Vertical Federated Learning (VFL) techniques, such as those proposed by Kang \textit{et al.} \cite{FEDCVT} and Wei et al. \cite{WeiVFL2022} attempt to address this challenge. However, they often require the sharing of intermediate representations, which may compromise privacy. Our approach differs by maintaining strict data isolation throughout the learning process.

Private Set Intersection (PSI) \cite{LuLinpeng2020} offers a potential solution for aligning data across silos. However, when extreme sample misalignment occurs, it leads to significant data reduction when applied across multiple silos. This data reduction problem is particularly significant in scenarios with a large number of silos.

Recent work by Qi et al. \cite{Qi2022VerticalFL} on vertical federated learning with contrastive learning shows promise for image data, but its applicability to tabular data remains unexplored. 

While the above methods have shown promising results in various vertical federated learning scenarios, they rely on data sharing or intermediate representation exchange that violates the strict privacy constraints of our target environment. Our work addresses a more restrictive setting where such data sharing is not permissible, necessitating a novel approach.

\subsection{Learning on  Data Silos with Non-IID}
The heterogeneity of data distribution can lead to misalignment within the context of federated learning \cite{dandi2021implicitgradientalignmentdistributed}.  One reason for this is the presence of non-IID data within the federated learning network. Hsieh et al. \cite{Hsieh2020} provide a comprehensive analysis of how non-IID data results in skewed distributions across different silos. Existing approaches to address this issue include...:
\begin{enumerate}
    \item Data sharing: Zhao et al. \cite{Zhao2018} proposed partial data sharing to mitigate non-IID effects, but this compromises privacy.
    \item Incentive mechanisms: Wang et al. \cite{Wang2022} explored using rewards to incentivize silos with good accuracy, but this approach introduces complexity and bias due to dimension reduction. The approach is also specific to image data.
    \item Transfer learning: Tzinis et al. \cite{Tzinis2021} demonstrated the effectiveness of transfer learning for non-IID speech data, but their method requires pre-trained models, which may not be available in all domains.
\end{enumerate}
Our CFL method takes a different approach by leveraging contrastive learning to create more robust representations that can better handle non-IID data without requiring data sharing or pre-trained models.

\subsection{Learning with label costliness }

Label scarcity due to label costliness is a common challenge in many real-world machine-learning applications, particularly in federated settings where labelling efforts may be distributed and inconsistent. Contrastive learning (CL) has shown promise in addressing this issue by learning useful representations from unlabeled data.

While CL has been successfully applied to image, text, and speech data \cite{Gutmann2010, Chen2020SimCLR}, its application to tabular data has been limited. LaaF \cite{Hager_2023_CVPR}, SubTab \cite{ucar2021subtab} and SCARF \cite{scarf} made progress in this direction by proposing partial data augmentation for tabular contrastive learning. However, their approach does not address the unique challenges of federated learning environments.

The work most closely related to ours is FedCVT \cite{FEDCVT} and MOON \cite{Li2021contrastiveFL}, which attempt to combine federated and contrastive learning. However, these methods still require either representation sharing or supervised learning during the training phase, which may not be feasible in a strict privacy restriction environment.

\subsection{Theoretical Foundations}
To provide a stronger theoretical basis for our work, it is important to note that different theoretical frameworks underpin federated learning and contrastive learning. Federated learning builds on distributed optimization theory \cite{smith2017federated, yang2019federated}, while contrastive learning is in information theory and representation learning \cite{tian2020contrastive}. Our CFL method aims to bridge these theoretical foundations and employs the global learning of FL with the representation learning capabilities of CL.


\section{Problem Formulation}
\subsection{Definition}
We consider a federated learning environment with $N$ data silos, each containing vertically partitioned tabular data. The complete feature set of data consists of all features from the silo space. This complete set of data is never collected together but connected through a quasi-identifier. The concept is presented to represent real-world environments and to bridge the gap between contrastive learning and federated learning.  We mentioned it as the global (imaginary)  data. The problem in our research is defined as follows:

\subsubsection{\textbf{Data distribution}}
Let $D_i=\{(x_i^j,y^j)\}_{j \in S_i}$ denote the dataset in the $i$-th silo, where:
    \begin{itemize}
        \item ${x_i^j \in R^{d_i}}$  is the feature vector for the $j$-th sample in a silo $i$
        \item $y^j  \in Y$ is the corresponding label (if available)
        \item   $S_i$ is the set of sample indices available in silo $i$
        \item $d_i$ is the feature dimension in a silo $i$

    \end{itemize}
The total dimension of the feature across all silos is $d=\sum_{i=1}^{N}d_i$.

\subsubsection{\textbf{Vertical partition}}
For each sample $j$, the complete feature vector is $x^j=[x_1^j;x_2^j;...;x_N^j] \in R^d$, where ";" denotes concatenation. However, not all samples exist in all silos.

\subsubsection{\textbf{Global (imaginary) data}}
For each sample $j$, the complete feature vector is $x_G^j=[x_1^j,x_2^j,...,x_N^j] \in R^d$ , $G$ is a notation of the global dataset. Therefore, $D_G=\{(x_G,y_G)\}$. Although not always available, in our silo spaces, apply $y_G=y_i$ where $i$ is a silo number.

\subsubsection{\textbf{Ordered index}}
During the training process, an ordered index $I=\{1,2,...,m\}$ is established across all silos, where $m$ is the number of unique samples across all silos. Each silo $i$ maintains a mapping $\phi_i:S_i\rightarrow I$ that associates its local samples to this common index. This allows for:
\begin{itemize}
    \item Consistent reference to samples across silos without revealing actual data
    \item Handling of missing data when a sample index exists in $I$ but not in $S_i$ for a particular silo
\end{itemize}

\subsubsection{\textbf{Non-IID nature}}
The data distribution $P_i(X_i,Y)$ varies between silos, that is, $P_i(X_i,Y)\neq P_j(X_j,Y)$ for $i\neq j$.

\subsubsection{\textbf{Label scarcity due to label costliness}}
In each silo $i$, only a fraction $\alpha_i\in(0,1]$ of samples are labelled, which typically $\alpha_i << 1$. We define:
    \begin{itemize}
        \item $D_i^L=\{(x_i^j,y^j)|y^j \text{ is known}\}$: the labelled subset
        \item $D_i^U=\{x_i^j|y^j \text{ is unknown}\}$ : the unlabelled subset 
        \item Such that $|D_i^L|= \alpha_i n_i$ and $|D_i^U|=(1 \text{-} \alpha_i)n_i$.

    \end{itemize}

\subsubsection{\textbf{Sample misalignment}}
Each silo $i$ has access to only a fraction of the total samples, and the samples across silos may partially overlap or be similar but not identical. Formally:
\begin{itemize}
    \item $|S_i|=\beta_{i}*n_{total}$, where $n_{total}$ is the total number of unique samples across all silos
    \item $S_i \neq S_j$ for $i \neq j$ in general
    \item The intersection of sample sets across silos may be non-empty: $\cap_{i=1}^N S_i\neq 0$

\end{itemize}

\subsubsection{\textbf{Linkage mechanism}}
Samples across silos are linked using quasi-identifiers $q^j$, such that $q_i^j=q_k^j$ for the same sample $j$ in different silos $i$ and $k$.

\subsubsection{\textbf{Strict privacy constraints}}
No raw data (features, feature names) can be shared between silos. Only derived information (e.g., model parameters, gradients) can be exchanged.

\subsubsection{\textbf{Contrastive learning}}
Contrastive learning generates new data with better-supervised learning performance. Let $C$ be the contrastive learning function, $E:R^d \rightarrow R^p$ be an encoder function that maps the input space to a $p$-dimensional embedding space and $D:R^p \rightarrow \hat{Y}$ be a decoder function that maps the embedding space to the predicted label. The contrastive learning objective is defined as:
\begin{itemize}
    \item During training $C:D(E) \rightarrow \hat{Y}$
    \item During inference $C:E \rightarrow R^p $
\end{itemize}

Our study addresses the issue of vertical partitioned data silos, focusing on learning with sample misalignment across silos without raw data sharing. In this study, we do not discuss topics related to secure index sharing and secure gradient exchange.
\subsection{Problem Statement}
Our study is to develop a Contrastive Federated Learning with Data Silos (CFL) method that addresses vertical partitioning with sample misalignment.

If a model that is trained with a global (imaginary) dataset is $f_G:R^d \rightarrow Y _G$ and a model trained locally from a silo is $f:R^{d_i} \rightarrow Y_i$ , then our objective is to create :
\begin{gather}
    f:R^{d_i} \rightarrow Y_i \approx  f_G:R^d \rightarrow Y_G
\end{gather}
Subject to:
\begin{itemize}
    \item $f$ is trained with a local (silo) dataset.
    \item No Raw Data Sharing: $   
    \forall i,j \in \{1,\ldots,N\} \text{ with } i \neq j : D_i \cap D_j = \emptyset
    $
Where $\{1,\ldots,N\}$ is an ordered index set, and $D_i$, $D_j$ are feature sets.
    \item Sample misalignment: $S_i \neq S_j$ and $\cap_{i=1}^N S_i\neq 0$ 
\end{itemize}

\section{Proposed Method}
We introduce Contrastive Federated Learning with Data Silo (CFL) as a solution to the problems. CFL combines tabular contrastive learning (CL) with unique features (zero-fill, tuple representation, Pearson reordering) to solve vertical federated learning without loosening privacy.
\subsection{Pre-processing Adaptation}
Our CFL uses zero fill and Pearson reordering to adapt to vertical federated learning and contrastive learning.

 \textbf{Zero fill for missing samples.} Our CFL adopts zero fill to ensure that the data are available in each silo.  This is introduced to solve sample misalignment. 
For each silo $i$, we create a complete dataset $D_i$ by zero-filling missing samples $D_i=\{(x_i^k,y^k)|k\in I\}$.  If $k \notin S_i$, we set $x_i^k = 0 $ (zero vector of dimension $d_i$)
In some of our experiment settings, we left only 25\% of the data in some silos. During local learning, there is a 75\% chance that the slice $x_i^k$ in $D_i'$ does not exist in  silo $i$. For example, suppose data with an object identifier $k=2$ exist in the first silo but are absent in the second silo during training. In that case, the second silo is trained with a zero matrix representing $i=2$ in silo 2. The matrix below was the result of the above example: $$ \mathcal{X} =  \begin{bmatrix} R^1_1\ \\ R^1_5 \\ R^1_2\\ R^1_6 \end{bmatrix} = \begin{bmatrix} a^1_1 & b^1_1 & c^1_1 & d^1_1 & e^1_1 \\ a^1_5 & b^1_5 & c^1_5 & d^1_5 & e^1_5 \\ \_ & \_ & \_ & \_ & \_ \\ a^1_6 & b^1_6 & c^1_6 & d^1_6 & e^1_6 \end{bmatrix} \Rightarrow \begin{bmatrix} a^1_1 & b^1_1 & c^1_1 & d^1_1 & e^1_1 \\ a^1_5 & b^1_5 & c^1_5 & d^1_5 & e^1_5 \\ .0 & .0 & .0 & .0 & .0 \\ a^1_6 & b^1_6 & c^1_6 & d^1_6 & e^1_6 \end{bmatrix} $$
Zero-fill is selected because it is fast and requires neither data calculation nor data sharing. Furthermore, during training, $D_i$ is added with noise locally and aggregated with FL to minimize zero imputation. 

Locally (within a silo), zero imputation may result in deviation in covariance. However, as the number of silos grows during FL, the deviation will be minimized toward zero.  Let the covariance deviation of silo $i$ be:
\begin{gather}
    \Vert {\Sigma}^{true}_i - {\Sigma}^{imp}_i  \Vert _F < \delta_i  
\end{gather}  
where $\Sigma^{true}$ is the covariance of true matrix,  $\Sigma^{imp}$ is the covariance of zero-fill matrix, $\Vert \cdot \Vert_{F}$ is the Frobenius norm, $\delta$ is the local deviation bound for silo $i$ , and the global covariance during federated learning are:
\begin{align}
    \Sigma_{global}^{true} &= \frac{1}{N} \sum_{i=1}^M n_i \Sigma_i^{true} \\
    \Sigma_{global}^{imp} &= \frac{1}{N} \sum_{i=1}^M n_i \Sigma_i^{imp} 
    \end{align}
where $n$ is number of sample in silo $i$ and $N=\sum_{i=1}^M n_i$, then the global deviation can be bounded:
\begin{align}
    \Vert \Sigma_{global}^{true} - \Sigma_{global}^{imp} \Vert _F &= \Vert \frac{1}{N} \sum_{i=1}^{M} n_i (\Sigma_i^{true} - \Sigma_i^{imp}) \Vert_F \label{eq:1} \\
    &\leq \frac{1}{N} \sum_{i=1}^M n_i \Vert \Sigma_i^{true} - \Sigma_i^{imp} \Vert_F \label{eq:2} \\
    & \leq \frac{1}{N} \sum_{i=1}^M n_i \delta_i  \label{eq:3}
\end{align}
Frobenius norm triangle inequality applies to Formula \ref{eq:1} and \ref{eq:2}. Matrix multiplicative property applies to Formula \ref{eq:3}. Because CFL is a vertical federated learning with sample misalignment problem, then $n_i \approx  \frac{N}{M}$ for all $i$, therefore:
\begin{align}
    \Vert \Sigma_{global}^{true} - \Sigma_{global}^{imp} \Vert _F & \leq \frac{1}{M} \sum_{i=1}^M  \delta_i
\end{align}
As M increases, the global covariance deviation converges to zero.
\begin{align}
    {\lim}_{M \rightarrow \infty} \Vert \Sigma _{global}^{true} - \Sigma _{global}^{imp} \Vert _F & \leq {\lim}_{M \rightarrow \infty} \frac{1}{M} \Sigma_{i=1}^M \delta_i \\
    & \leq {\lim}_{M \rightarrow \infty} \frac{\Delta}{M} = 0
\end{align}
Where $\Delta$ is some constant that bound $\delta_i \leq \Delta$.

\begin{figure}[H]
    \centering
    \includegraphics[width=0.9\linewidth]{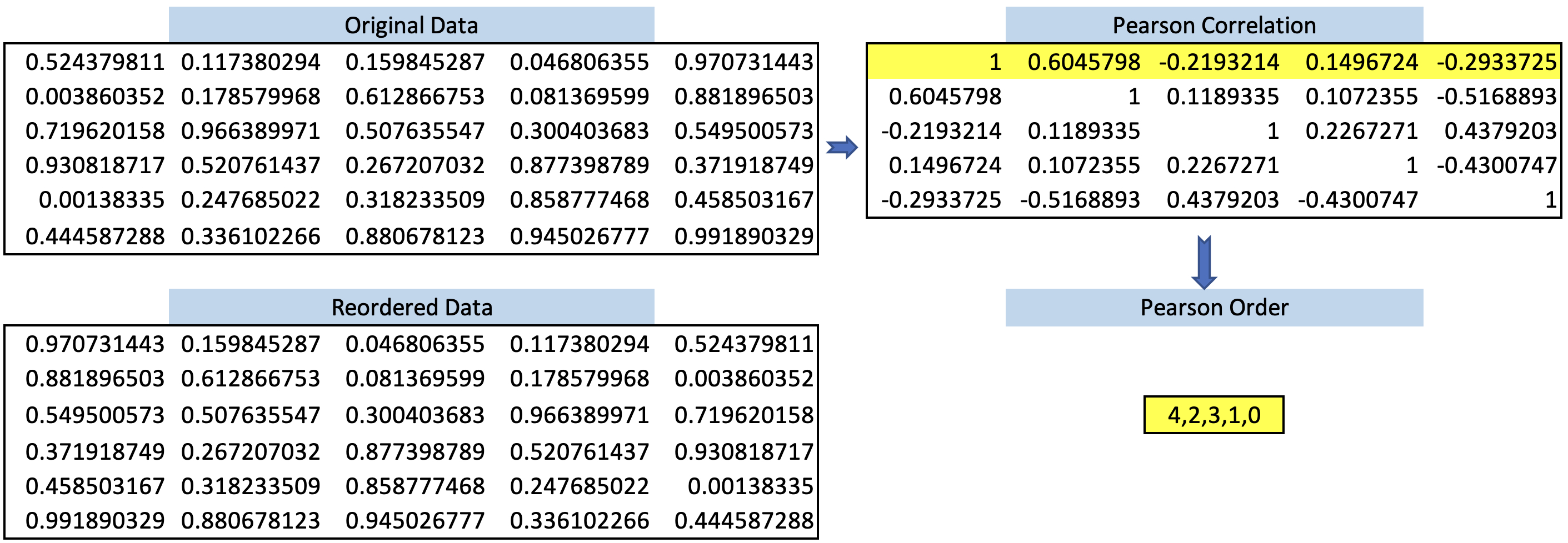}
        \caption{Our Pearson Ordering Processes. The original data are ordered by their Pearson correlation value to get a semantic representation useful for contrastive learning. This is to get a horizontal semantic relationship}
        \label{fig:PearsonProcess}
\end{figure}

\textbf{Pearson reordering for contextual transformation.}
To adapt tabular data into contrastive learning, CFL uses Pearson reordering. CL is an image/text-based algorithm that has spatial or contextual relation properties. CFL use Pearson correlation \cite{sedgwick2012pearson} to obtain this.
Figure \ref{fig:PearsonProcess} illustrate the process of the Pearson reordering. Let $P$ be the Pearson correlation and $\mathcal{D}_i^P$ be the sorted/ordered data; then $\mathcal{D}_i^P = \textit{sort}(P({\mathcal{D}_i}))$ is achieved by sorting the Pearson correlation value. This was done on the assumption that we never knew the actual contextual order of the data. In our CFL, $\mathcal{D}_i$ is always $D_i^P$. Therefore, we always refer to it as $D_i$.
\subsection{CL in our CFL}
\begin{figure}[H]
    \centering
    \includegraphics[width=1\linewidth]{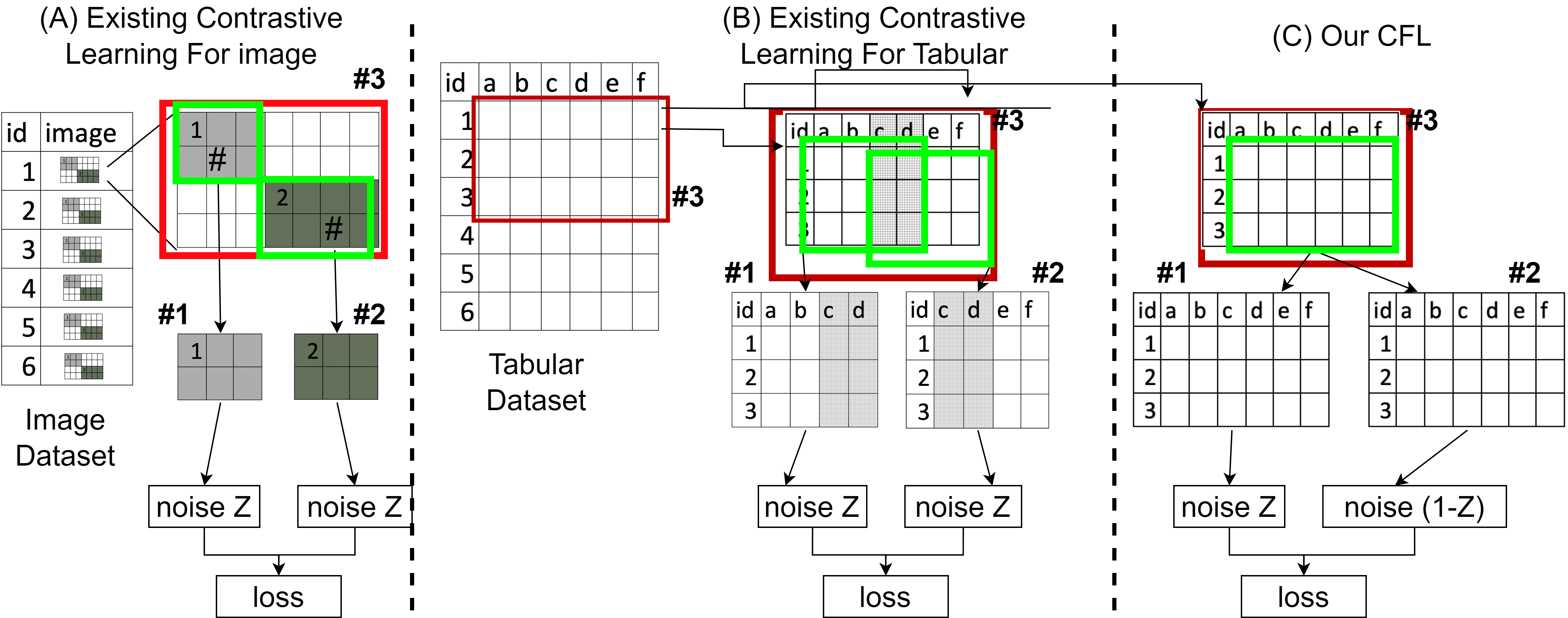}
        \caption{ \{a,b,c,d,e,f\} is the column name on the tabular data, (\#1) is the representation $1^{st}$, (\#2) is the representation $2^{nd}$, and (\#3) is a set of data targeted for the loss calculation. In (A), the representations are generated from a \textbf{single} record (\#3) (single ID ). In (B) and (C), the representations are generated from a \textbf{set} of records (\#3) (several IDs). In (B), the representations are built from part of the data (\#3) with some intersection (dark area in B), $\{\#1 \subseteq \#3,\#2 \subseteq \#3, \#1 \cap \#2\}$. In our CFL (C), each representation is a clone of the data (\#3), $\{\#3 = \#1 = \#2\}$ / full-row representation.}
        \label{fig:augmentation}
\end{figure}

To adapt contrastive learning to the silo / vertical federated learning problem, we implement CL with tuple representation in our CFL, as demonstrated in Figure \ref{fig:augmentation}. This approach is closely aligned with the principles of contrastive learning for image data outlined by Yao et al. \cite{Yao2022}. 
As shown in the Figure \ref{abstract}, we use a full representation because the data available within the context of our study is actually part of global (imaginary) data (a slice). Therefore, during contrastive learning, CFL does not slice the data anymore as the join (conceptually) will be done within federated learning on the global server.
Our CFL first replicates the data to create the representation required for contrastive learning. Let a slice of data be used in a local contrastive learning step  $\mathcal{B} = [b] \text{ where }  \mathcal{B} \subseteq \mathcal{D}_i$. $\mathcal{B}$ is cloned into two objects $\mathcal{B} \rightarrow \{\mathcal{B}^1, \mathcal{B}^2\}$.  Second, $\mathcal{B}^1$ and $\mathcal{B}^2$ are subject to a binomial mask and additional modifications, such as swapping or introducing Gaussian noise. The noise rate is $(z_1,z_2)$ for each pair. 
Third, let $X=[x]$ be noisy $\mathcal{B}$,  then $([x^1],[x^2])$  inputted into the contrastive learning encoder and decoder layers. 
If $\bar{E}:x;\omega^e \rightarrow x^e$  be an encoder function given the parameter $\omega^e$ and $\bar{D}:x^e;w^d \rightarrow x^d$, be a decoder function given $\omega^d$ parameter,  the contrastive learning function $f_c$ can be  written as:
\begin{gather}
     f_c:\bar D: (\bar E :x;\omega^e);\omega^d \rightarrow x^d
\end{gather}
Within contrastive learning, for each $x$, the total loss $L_t$ is calculated as follows:
 \begin{equation}
\label{eq:totalLossFunction}
 L_t(x)^{(1,2)} = L_r(x)^{(1,2)}+ L_c(x)^{(1,2)} + L_d(x)^{(1,2)}
\end{equation}
Where  $L_r$ is the reconstruction loss, $L_c$ is the contrastive loss, and $L_d$ is the distance loss.
The objective of contrastive learning is to minimize the total loss $L_t$.  
\begin{equation}
\label{eq:totalLossFunction2}
\begin{split}
 \text{arg min } L_{tn}(\mathcal{X};\omega^e,\omega^d) &= \text{arg min }  \frac{1}{J} \Sigma_{j=1} ^{J} L_t(x)^{1,2} 
\end{split}
\end{equation}
When $MSE(.)$ is the mean square error function then:
\begin{equation}
\label{eq:reconstruction}
\begin{split}
L_r(x) &= \dfrac{1}{N} \sum_n^N MSE(x^d,b) \\
L_d(x) &= \dfrac{1}{N} \sum_n^N MSE(x^e)^{(1,2)}
\end{split}
\end{equation}
Compared to SubTab, which uses a similar function, such as cosine distance, we simplify contrastive loss $L_c$ with only a result of a dot product.  
\begin{equation}
\label{eq:lossC}
\begin{split}
L_c(x) &= \frac{1}{N} \sum_{n}^N l(x^e)^{(1,2)} \\
L_c(x) &= \frac{1}{N} \sum_{n}^N (-\log \frac{\exp(MSE([0],dot(x^{e1} , x^{e2}) \text{ / } \mathcal{T}))}{\sum_{k=1}^K \exp(MSE([0],dot(x^{e1} , x^{e2}) \text{ / } \mathcal{T}))})
\end{split}
\end{equation}

This modification leads to a faster model. This is because in modern architecture, a modern processing unit (CPU/GPU) uses an optimized parallel algorithm for the operation of the dot product \cite{blas}.  We tested this in one of our experiments. The pseudocode for data generation can be found in Algorithm \ref{alg:noisiData}. 

\begin{algorithm}
\caption{Generate Noisy Representations $(x_i,x_j)$}
\label{alg:noisiData}
\begin{algorithmic}[1]
\Require
    \State $x$ \Comment{Original tabular data sample}
    \State $\sigma$ \Comment{Noise level}
    \State $p$ \Comment{Probability of feature masking}
\Ensure
    \State $x_i, x_j$ \Comment{Two noisy representations of $x$}

\Function{GenerateNoisyRepresentations}{$x, \sigma, p$}
    \State $x_i \gets x$
    \State $x_j \gets x$
    
     \State $x_i$ = BinomialMask($x_i$,$\text{noiseRate } \sigma$)
    \State $x_j$ = BinomialMask($x_j$, $\text{noiseRate } \sigma $) + GausianNoise($x_j$)
    
    \Return $x_i, x_j$
\EndFunction

\Procedure{Main}{}
    \State $x \gets \text{LoadTabularSample}()$
    \State $\sigma \gets 0.1$ \Comment{Adjust noise level as needed}
    \State $p \gets 0.2$ \Comment{Adjust masking probability as needed}
    
    \State $x_i, x_j \gets \text{GenerateNoisyRepresentations}(x, \sigma, p)$
    \State \text{UseForContrastiveLearning}($x_i, x_j$)
\EndProcedure
\end{algorithmic}
\end{algorithm}

\subsection{FL in our CFL}
CFL merge the CL parameters from each silo with FL to learn global knowledge. Both decoder and encoder parameters $(\omega^e,\omega^d)$ from local CL are aggregated within a global server. The aggregation was carried out by averaging the parameters of each silo with a federated average (FedAVG) denoted as :
\begin{gather}
    F(g) = \dfrac{1}{N} \sum_{n=1}^N (\omega^e,\omega^d) \rightarrow (\omega^{eG},\omega^{dG})
\end{gather}
Where $(\omega^{eG},\omega^{dG})$ are the global averaged parameters. 
The aggregated global parameters are then returned to each silo for back-propagation operations to continue contrastive learning.

At the end of the learning loop, the header (decoder) is omitted. Therefore, during supervised learning in each silo, the functions were originally $ f:\mathcal{D}_i \rightarrow Y_i$ by performing contrastive learning, it becomes $f:  E : \mathcal{D}_i;\omega^{eG} \rightarrow Y_i$.  Our goals mentioned before:
\begin{equation}
    \begin{split}
        f_G:R^d \rightarrow Y_G &\approx f:R^{d_i} \rightarrow Y_i\\
        \text{due to contrastive learning :}
         f_G:D &\approx f:  E:\mathcal{D}_i;\omega^{eG} \\
         \text{where } f
         :  E:\mathcal{D}_i;\omega^{eG} &> f:R^{d_i} \rightarrow Y_i
    \end{split}
\end{equation}
Note that $\mathcal{D} = R^d$they are written like above to connect with previous definitions. By employing $F(g)$, CFL maintains strict privacy constraints. Neither original nor representation data are shared, only parameters.
The pseudocode of CFL can be found on Algorithm \ref{alg:cfl}.
\begin{algorithm}
\caption{Contrastive Federated Learning with Data Silos (CFL)}
    \label{alg:cfl}
\begin{algorithmic}[1]
\Require
    \State $\{D_1, D_2, ..., D_N\}$ \Comment{Datasets for $N$ clients}
    \State $T$ \Comment{Number of communication rounds}
    \State $E$ \Comment{Number of local epochs}
    \State $\eta$ \Comment{Learning rate}
    \State $\tau$ \Comment{Temperature for contrastive loss}
\Ensure
    \State $\theta_g = (\omega^{eG},\omega^{dG}) $  \Comment{Global model parameters}

\Procedure{ServerUpdate}{$\{\theta_1, \theta_2, ..., \theta_N\}$}

    \State $\theta_g \gets \frac{1}{N} \sum_{i=1}^N \theta_i$ \Comment{Aggregate client models}
    \Return $\theta_g$
\EndProcedure

\Procedure{ClientUpdate}{$\theta_g, D_i$}
    \State $\theta_i \gets \theta_g$ \Comment{Initialize local model with global parameters}
        \For{batch $b$ in $D_i$}
            \State $b_1, b_2 \gets \text{Augment}(b)$ \Comment{Generate two augmented views}
            \State $X_1, X_2 \gets \text{GenerateNoisyRepresentations}(b_1, b_2 )$ \Comment{Generate two Noisy views}
            \State $E_{1,2} \gets \text{Encoder}(X_{1,2})$ \Comment{Encode views}
            \State $Z_{1,2} \gets  \text{Dencoder}(E_{1,2})$ \Comment{Dencode views}
            \State $\mathcal{L}_{\text{dist}} \gets \text{Distance}(Z_1, Z_2, \tau)$ \Comment{Compute distance loss}
            \State $\mathcal{L}_{\text{con}} \gets \text{ContrastiveLoss}(Z_1, Z_2, \tau)$ \Comment{Compute contrastive loss}
            \State $\mathcal{L}_{\text{rec}} \gets \text{ReconstructionLoss}(X_1, X_2, Z_1, Z_2)$ \Comment{Compute reconstruction loss}
            \State $\mathcal{L} \gets \mathcal{L}_{\text{con}} + \mathcal{L}_{\text{rec}} + \mathcal{L}_{\text{dist}}$ \Comment{Total losses}
            \State $\theta_i \gets \theta_i - \eta \nabla \mathcal{L}$ \Comment{Update local model}
        \EndFor
    \Return $\theta_i =(\omega^{e},\omega^{d})$
\EndProcedure

\Procedure{CFL}{$\{D_1, D_2, ..., D_N\}, T, E, \eta, \tau$}
    \State Initialize $\theta_g$
    \For{$e \gets 1$ to $E$}
        \For{each client $i$ in parallel}
            \State $\theta_i \gets \text{ClientUpdate}(\theta_g, D_i)$
        \EndFor
        \State $\theta_g \gets \text{ServerUpdate}(\{\theta_1, \theta_2, ..., \theta_N\})$
    \EndFor
    \Return $\theta_g = (\omega^{eG},\omega^{dG})$
\EndProcedure
\end{algorithmic}
\end{algorithm}

\section{Experiments}
We conducted extensive experiments to evaluate the effectiveness of our proposed Contrastive Federated Learning with Tabular Data Silos (CFL) method. Our experiments involved six datasets and four different experimental settings to simulate various real-world scenarios.
\subsection{Datasets}
The experiments in this study involved six datasets: the Adult Income dataset \cite{misc_adult_2} (income), the BlogFeedback dataset \cite{misc_blogfeedback_304} (blog), the synthetic biometric blender dataset \cite{STIPPINGER2023101366} (syn), the Sensorless Drive Diagnosis dataset \cite{misc_dataset_for_sensorless_drive_diagnosis_325} (sensorless/Sls), the Covertype dataset \cite{misc_covertype_31} (covtype), and the TUANDROMD (Tezpur University Android Malware Dataset / Tumod) \cite{misc_tuandromd_(tezpur_university_android_malware_dataset)_855}. 

The Biometric Blender Synthetic Dataset (BB) was chosen because of its ability to generate multiple features. BB dataset is used to create a dataset consisting of 10 classes, with each row containing 1600 features. The complete setup for every experiment conducted in this work is shown in Table \ref{tab:experimentSetup}. 
\begin{table}
\small
    \centering
\caption{Experiments Setup. Six datasets, e.g. Cover Type (Covtype), Blog (Blog), Adult (Adult), Biometric Synthetic (Syn), Sensorless Drive Diagnosis (Sls), TUANDROMOD (Tumod) are used .}
\label{tab:experimentSetup}
    \begin{tabular}{|nn|rrrrrr|}
    \toprule
         \multicolumn{2}{|l|}{Dataset}&  Covtype&  Blog&  Adult&  Syn&Sls& Tumod\\
         \midrule
         \multicolumn{2}{|l|}{Rows}&  581012&  52396&  30162&  999000&  58509& 4464\\
         \multicolumn{2}{|l|}{Full Features}&  54&  280&  105&  1600&  49& 241\\
         \midrule
         Standar&  Silos Count&  3&  4&  5&  4&  4& 6\\
         &  Encoder Size&  256&  256&  256&  2048&  256& 256\\
         &  Feature Size&  18&  70&  21&  400&  12& 40\\
         \midrule
         Data Drop&  Silos Count&  3&  4&  5&  4&  4& 6\\
         &  Encoder Size&  256&  256&  256&  2048&  256& 256\\
         &  Feature Size&  18&  70&  21&  400&  12& 40\\
         \midrule
 Class Imbalance& Silos Count& 3& 4& 5& 4& 4&6\\
 & Encoder Size& 256& 256& 256& 2048& 256&256\\
         &  Feature Size&  18&  70&  21&  400&  12& 40\\
         \midrule
 Mixed& Silos Count& 3& 4& 5& 16& 4&6\\
 & Encoder Size& 256& 256& 256& 2048& 256&256\\
 & Feature Size& 18& 70& 21& 100& 12&40\\ \hline
    \end{tabular}

\end{table}

\begin{table}
\caption{Information of model used in this experiment's evaluations}
\small
\centering
\begin{tabular}{|l|L|l|}
\hline
Model Name & Information                                                                    & Function \\ \hline
Base 1     & Logistic regression results on the global (imaginary) data                       & $f:R^d$\\ \hline
CFL (Ours) & Logistic regression results on CFL on local data with data silo learning       & $f: E:\mathcal{D};\omega^e$\\ \hline
Base 2     & Logistic regression results on local data                                      & $f:R^{d_i}$\\ \hline
SubTab     & Logistic regression results on SubTab on local data without data silo learning & $f_{lc}: E\mathcal : {D};\omega^e$\\ \hline
SubTab FL  & Logistic regression results on SubTab on local data with data silo learning    & $f_{fl}:  E : \mathcal{D};\omega^e$\\ \hline
\end{tabular}
\label{infoModel}
\end{table}

\subsection{Experiment Settings}
Due to the unique privacy requirements of our study, direct comparisons with existing federated learning methods are not possible. Instead, we compare our method against baselines that respect these strict privacy constraints. We compared our CFL method against 4 model baselines, e.g. 
Base 1, 
Base 2, 
SubTab,
SubTab FL, see FIgure \ref{infoModel}.  We also compare CFL with existing deep learning networks such as MLP, Scarf, and Transformer.

We designed four experimental settings to evaluate our CFL method under different conditions:
\begin{enumerate}
    \item \textbf{Standard Lab Setting}.
This setting ensures data availability and avoids sample misalignment. It serves as a baseline for understanding the models' performance.
    \item \textbf{Data Size Imbalance Setting}.
We introduced an unequal distribution of data sizes within the silo space to simulate small sample scenarios, which will lead to sample misalignment. The experiment included a client dropout rate of 0.25 and a data dropout rate of 0.5 for most datasets, except for the cover-type dataset, which had a client dropout rate of 30\%. In a scenario where 25\% client drop rate is applied, and there are a total of 4 clients, the first client would encounter a data size imbalance. If \( N \) is the total number of clients and $D $ is the total data available, then the data owned by a client without imbalance is  $D$. For a client with a data size imbalance, the data received is \( \frac{1}{2} \times D \). The complete configuration is detailed in Table \ref{tab:experimentSetup}.
\begin{algorithm}
\caption{Generate Data for Sample Misalignment Experiment}
\begin{algorithmic}[1]
\Require
    \State $D$ \Comment{Original dataset}
    \State $N$ \Comment{Number of clients/silos}
    \State $c_d$ \Comment{Client dropout rate}
    \State $d_d$ \Comment{Data dropout rate}
    \State $l_i$ \Comment{Class dropout rate}
\Ensure
    \State $\{D_1, D_2, ..., D_N\}$ \Comment{Datasets for each client}

\Function{GenerateImbalancedDataSize}{$D, N, c_d, d_d$}
    \State $C \gets \lfloor N \cdot c_d \rfloor$ \Comment{Number of clients with imbalance}
    \State $\{D_1, D_2, ..., D_N\} \gets \text{SplitDataEvenly}(D, N)$ \Comment{Initial even split}
    
    \For{$i \gets 1$ to $C$}
        \State $D_i \gets \text{RandomSample}(D_i, (1 - d_d) \cdot |D_i|)$ \Comment{Apply data dropout}
    \EndFor
    
    \Return $\{D_1, D_2, ..., D_N\}$
\EndFunction

\Function{GenerateImbalancedClassSize}{$D, N, c_i, l_i$}
    \State $C \gets \lfloor N \cdot c_i \rfloor$ \Comment{Number of clients with class imbalance}
    \State $\{D_1, D_2, ..., D_N\} \gets \text{SplitDataEvenly}(D, N)$ \Comment{Initial even split}
    \State $L \gets \text{UniqueLabels}(D)$ \Comment{Get all unique labels}
    \State $L_r \gets \lfloor |L| \cdot (1 - l_i) \rfloor$ \Comment{Number of labels to retain}
    
    \For{$i \gets 1$ to $C$}
        \State $L_s \gets \text{RandomSample}(L, L_r)$ \Comment{Randomly select labels to keep}
        \State $D_i \gets \text{FilterByLabels}(D_i, L_s)$ \Comment{Keep only data with selected labels}
    \EndFor
    
    \Return $\{D_1, D_2, ..., D_N\}$
\EndFunction

\Procedure{Main}{}
    \State $D \gets \text{LoadDataset}()$
    \State $N \gets 4$ \Comment{Number of clients, adjust as needed}
    \State $c_d \gets 0.25$ \Comment{25\% client dropout rate}
    \State $d_d \gets 0.5$ \Comment{50\% data dropout rate}
    \If{Data Size Imbalance}
    \State $\text{ImbalancedData} \gets \text{GenerateImbalancedDataSize}(D, N, c_d, d_d)$
    \EndIf
    \If{Class Size Imbalance}
    \State $\text{ImbalancedData} \gets \text{GenerateImbalancedClassSize}(D, N, l_i, d_d)$
    \EndIf

    \State \text{UseForExperiment}(ImbalancedData)
\EndProcedure
\end{algorithmic}
\end{algorithm}

    \item \textbf{Class Size Imbalance Setting}.
We introduced an imbalance in the distribution of data within the silo space with label/class size imbalance. We follow a study from Li \textit{et al.} \cite{li2021federatedlearningnoniiddata} , which proposes label imbalance causes non-IID data within federated learning. The experiment was designed with a 0.25 client imbalance and a 0.5 class imbalance for most datasets, except for the cover-type dataset, where there was a 30\% class imbalance. If $N=4$ is the total number of clients, the client with data size imbalance $C = \frac{1}{4} $ = 1, which means there is a client has class size imbalance. If $D$ is the total data available, then the data received by a client without imbalance is $D $. If $l^c=8$ is the unique number of labels, then clients without class size imbalance have $l^c=8$. For clients with class size imbalance $l^c=4$ , the label is randomly selected.  The complete setup is provided in Table \ref{tab:experimentSetup}.
\item \textbf{Mixed Case Settings}.
This setting combines both data size and class imbalances, representing the most challenging and realistic scenario.
\end{enumerate}

The experiment was conducted in an environment with extreme settings. We apply $0.3 : 0.7$ the training test rate.  However, due to sample misalignment, the training data could not be set to $0.1$ as typically found in contrastive learning experiments. For example, in some experiments, a drop rate of $0.5$ was applied to a client, which meant that the intended client had training data $D*(0.3*0.5)=D*0.15$. For the same reason, we skip the evaluation during training. 

The data were shuffled at each epoch, but the availability of specific data remained static across epochs. If a data point did not exist in the first epoch, it did not exist in other epochs.  The final data resulting from the CFL are applied to logistic regression \cite{kleinbaum2002logistic} to compare the results of each experiment and obtain evaluation scores.
\subsection{Additional experiments}
We conduct experiments to gain a better understanding of CFL.
\begin{enumerate}
    \item \textbf{Pearson reordering effect}.
    To evaluate the impact of our Pearson reordering approach, we conducted experiments in the standard setting with and without Pearson reordering. This allows us to isolate the effect of this preprocessing step on the overall performance of CFL. The evaluation is given with each dataset's mean of precision, recall and F1 scores.
    \item \textbf{Dot product loss effect.}
    We compared the performance of CFL using our simplified dot product loss function against the original contrastive similarity loss (e.g. cosine). This experiment was conducted in the standard setting to assess the impact of our loss function modification on computational efficiency. We target the time evaluation in seconds.
    \item \textbf{Comparison with Deep Learning Algorithms}.
We compare CFL model with other deep learning models to provide a more comprehensive analysis. Specifically, we use SCARF \cite{scarf}, MLP \cite{tabr} and Transformer \cite{tabr} . In addition, since CFL employs a contrastive learning algorithm, it can utilize any prediction head during inference. To enhance performance, we also integrate LightGBM \cite{ke2017lightgbm} into our CFL framework. 
\end{enumerate}

\subsection{Evaluation Metrics}
For each of the above experiments, we provide precision, recall, F1 (weighted), and delta, as shown in Table \ref{apdx}. Delta is the difference in F1 scores between the model when predicting data in each silo $f:R^{d_i}$ and the model when used with the global (imaginary)  data set $f:R^d$.

We present a summary table for each experiment, displaying the mean F1 scores and their deltas. The mean F1 score is calculated from the F1 scores of each model per dataset. The delta of the F1 score is denoted as:
\begin{gather}
\mathcal{A} : f: R^{d_i} \overset{\Delta}{=} \mathcal{A} :f:R^d    
\end{gather}
 
or in the context of contrastive learning, it is expressed as:
\begin{gather}
\mathcal{A} :f: \bar E(\mathcal{D};\omega^{eG}) \overset{\Delta}{=} \mathcal{A} :f:R^d    
\end{gather}

A lower value indicates closer proximity to the actual performance of the model with global data. For example, a value of -0.01 in the graph means:
\begin{gather}
    (\mathcal{A}( f: \bar E(R^{d_i};\omega^{eG})) - \mathcal{A}(f:R^d))= -0.01
\end{gather}

Where $\mathcal{A}(.)$ represents the performance function of a model. We define $\mathcal{A}(.)$ as F1 value score.
In cases of a positive value, it suggests that $\mathcal{A}(f:R^d) < \mathcal{A}( f: \bar E(R^{d_i};\omega^e))$.  

A bar graph and line graph are provided for a better understanding of each result. 
The bar graph illustrates the mean F1 score of the five models. 
The line graph represents the kernel density estimate (KDE) of the F1 scores used to assess the performance of the model within silos.
In the KDE graph, we anticipate observing a distribution with a shorter tail, indicating consistent model performance across all segments, regardless of the data.

\section{Result and Evaluation}
Key findings of  experiments:
\begin{enumerate}
    \item Consistent Performance: CFL consistently outperformed local models across all experimental settings, demonstrating its robustness in handling various data challenges.
    \item Global Model Competitiveness: In many cases, CFL matched or even surpassed the performance of models trained on global (imaginary) datasets, highlighting its effectiveness in federated settings.
    \item Imbalance Handling: CFL showed particular strength in scenarios with both data size and class imbalances, addressing a key challenge in real-world federated learning applications.
    \item Recall Improvement: While CFL improved both precision and recall, the gains in recall were particularly significant, indicating an enhanced ability to identify positive instances.
    \item Silo Performance Consistency: CFL demonstrated more consistent performance across different silos compared to other models, a crucial characteristic for federated learning systems.
    \item Synthetic Data Challenge: The synthetic (syn) dataset posed challenges for CFL in some settings, suggesting areas for potential improvement in handling certain types of artificially generated data.
    \item Pearson Reordering Impact: The introduction of Pearson reordering significantly enhanced CFL's performance across all datasets, particularly for income, covtype, and sensorless datasets.
    \item Efficiency Gain: The use of dot product loss instead of cosine similarity consistently reduced processing times, improving CFL's computational efficiency.
    \item Scalability: In the mixed case setting with an increased number of silos (16 for the syn dataset), CFL maintained its performance advantage, indicating good scalability.
    \item Privacy Preservation: CFL achieved these performance improvements while adhering to strict privacy constraints, neither sharing raw data nor intermediate representations
\end{enumerate}
\subsection{Sample Misalignment Across Silos}
\begin{figure}[]
    \centering
    \subfloat[CovType Dataset's Label Distribution]{
        \label{imgcovtype}
        \includegraphics[width=0.5\textwidth]{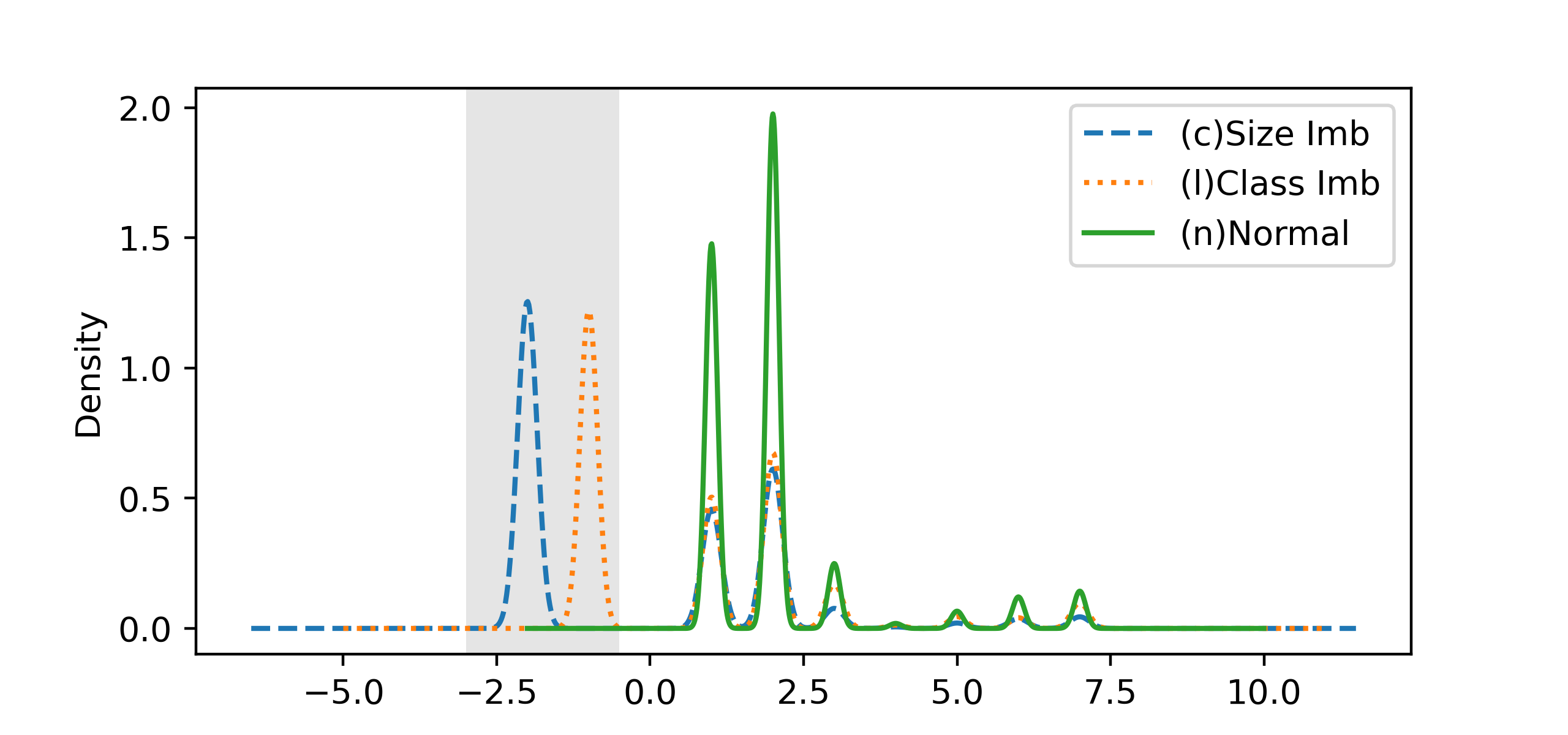}
    }
    \subfloat[Blog Dataset's Label Distribution]{
        \label{imgBlog}
        \includegraphics[width=0.5\textwidth]{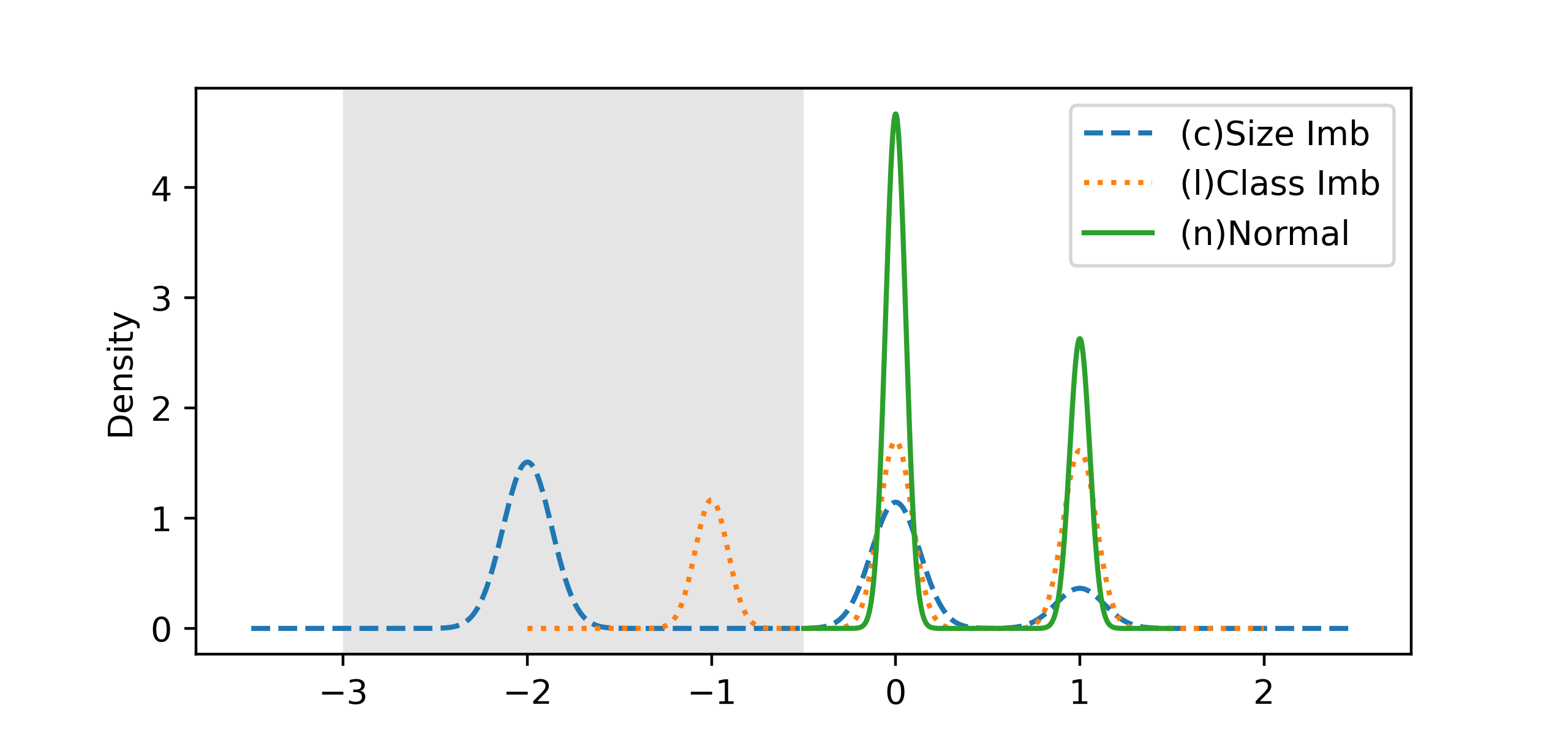}
    }
    \newline
    \subfloat[Adult Income Dataset's Label Distribution]{
        \label{imgAdult}
        \includegraphics[width=0.5\textwidth]{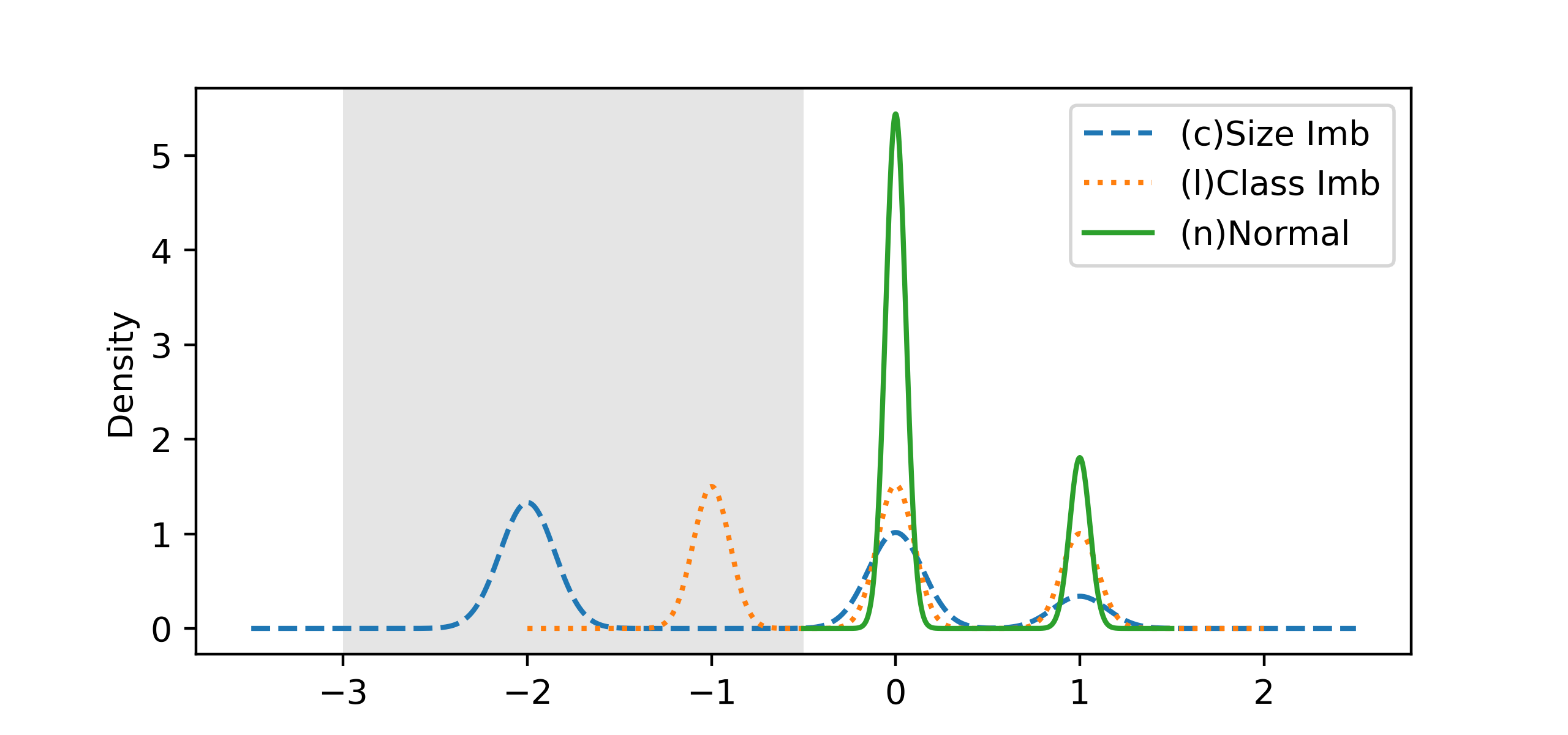}
    }
    \subfloat[Synthetic Dataset's Label Distribution]{
        \label{imgSyn}
        \includegraphics[width=0.5\textwidth]{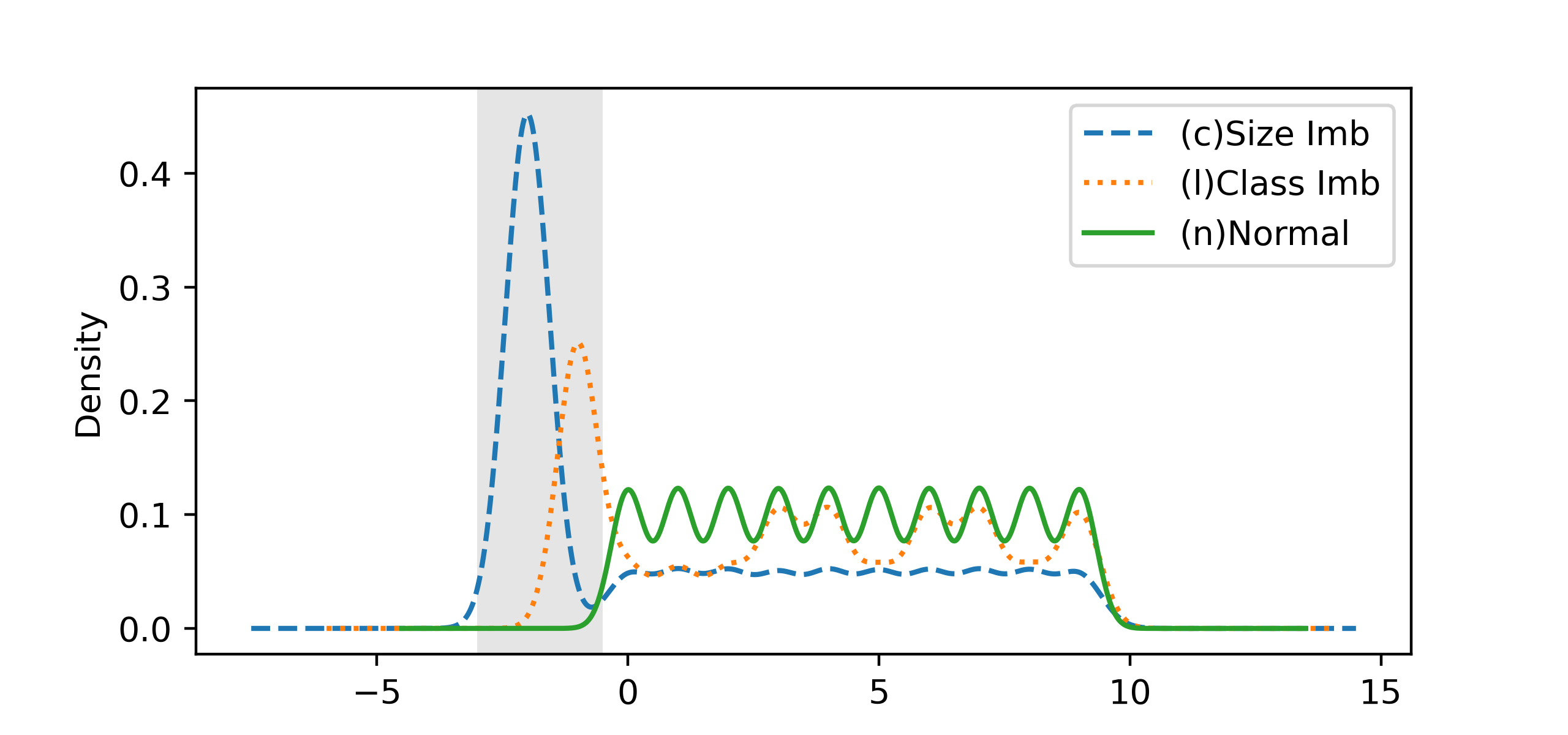}
    }
    \newline
    \subfloat[Sensorles Dataset's Label Distribution]{
        \label{imgSen}
        \includegraphics[width=0.5\textwidth]{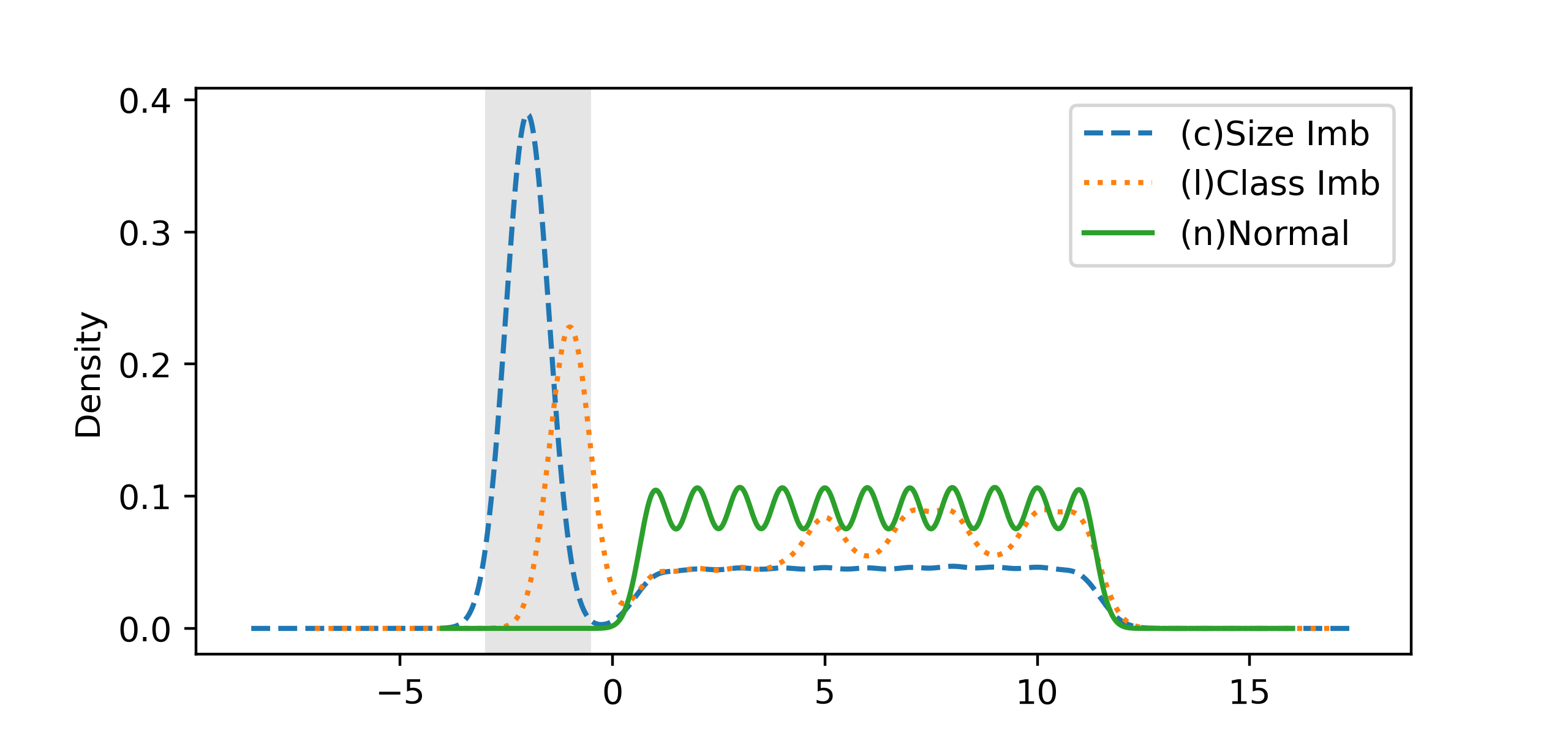}
    }
    \subfloat[Tuandromd Dataset's Label Distribution]{
        \label{imgtuandromd}
        \includegraphics[width=0.5\textwidth]{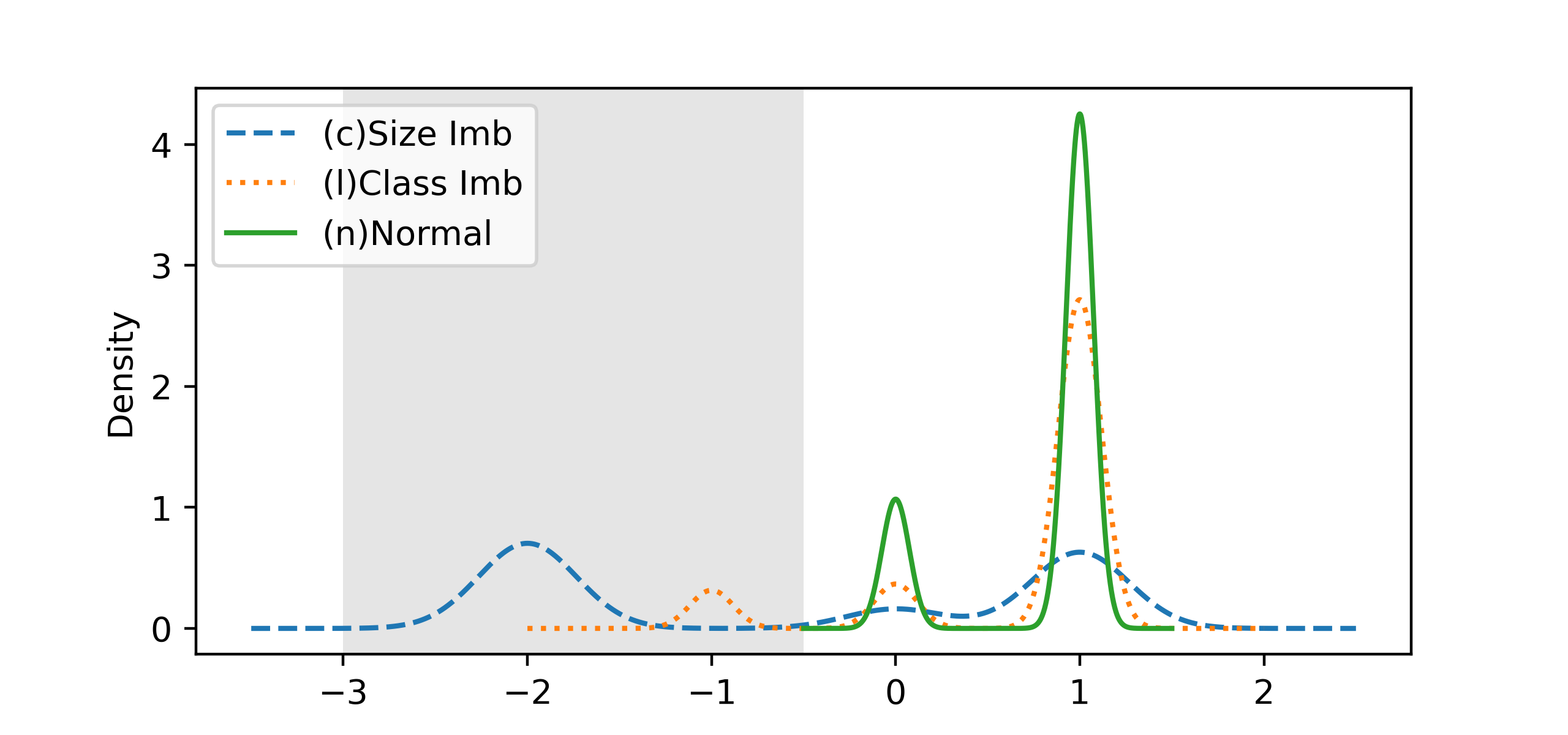}
    }
    \caption{Figures of data imbalance across silos. A value of -1 indicates dropped data in a client due to class size imbalance, while a value of -2 indicates dropped data in a client due to data size imbalance. (- -) or ($c$) represents a client with a data size imbalance, (. .) or ($l$) represents a client with a class size imbalance. Both are introduced to represent sample misalignment due to label costliness and non-IID within the data silo. The (--) or ($n$) represents a client without problems.}
    \label{allDataImabalance}
\end{figure}

Figure \ref{allDataImabalance} shows the sample misalignment visualizations. The figure shows how the data is distributed in three different client settings for each dataset, e.g., client with data size imbalance $c$, client with class size imbalance $l$, and client without data imbalance (normal) $n$. 
The line graph on value (-1) shows the number of data that are dropped based on random sampling to mimic a small sample from a silo. The line graph on value (-2) shows the number of data that are dropped based on class random sampling to mimic label costliness. The rest of the values are the true label. Data  $c,l,n $ are not evenly distributed.

From the figure, the normal client has more data compared to the other two clients. In clients with data imbalance, nearly half of the data is dropped during training. The data in each class is consistent, like in normal clients, but reduced by half. In a client with a class size imbalance, the data in each class is not uniform across silos. 

\subsection{Standard Setting}
Experiments conducted in a standard setting show that our proposed CFL model demonstrates the highest performance across all datasets except for the syn dataset, which is closest to the performance of the (imaginary) global model $f:D$. Table \ref{tab:subtab} presents a detailed performance of each model in each silo. From the delta data in Table \ref{tab:subtabResume}, it is clear that our CFL model improves the performance of the local model trained with local data and is comparable to the performance of the model with the global (imaginary) dataset. In particular, in the sensorless dataset, our CFL model shows superior performance, surpassing other models more than five times.

\begin{figure}[h]
    \centering
    \begin{minipage}{0.47\linewidth}
        \centering
        \includegraphics[width=1\linewidth]{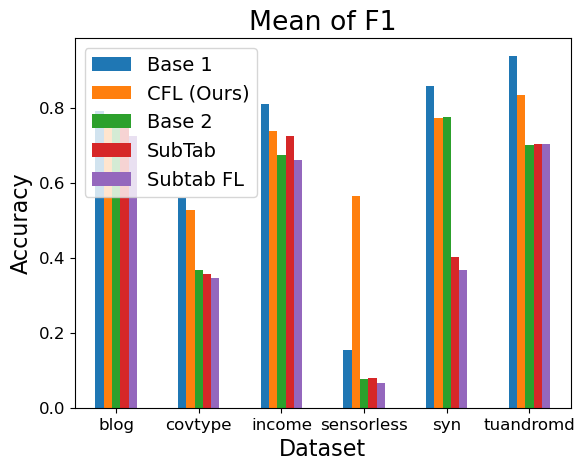} 
        \caption{Bar graphs depicting the average of F1 scores under standard conditions. In most instances, CFL outperforms other models. However, there is an exception in the syn dataset where CFL does not enhance the performance of the Base 2 (local) model.}
        \label{img:normalMean}
    \end{minipage}\hfill
    \begin{minipage}{0.47\linewidth}
        \centering
        \includegraphics[width=1\linewidth]{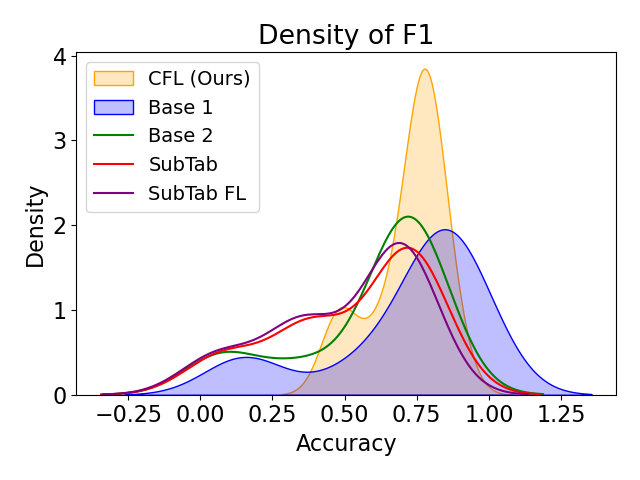} 
        \caption{Our CFL exhibits a greater Kernel Density Estimate (KDE) of F1 scores in comparison to other models, with a reduced tail. This indicates that the CFL has less variability in F1 scores across different silos.}
        \label{img:normalDensity}
    \end{minipage}
\end{figure}

Applying FL (Federated Learning) to SubTab decreases performance, which is expected since the original SubTab does not support FL.  SubTab without FL learning  performs worse than local logistic regression $f: \mathcal{D}_i$ in this setting. Figure \ref{img:normalMean} illustrates the average performance of each model. In Figure \ref{img:normalDensity}, our CFL model shows a shorter tail, indicating that performance in each silo is similar despite data variations. This aligns with the expected behaviour of the models derived from federated learning. Conversely, other models, such as Base 1, exhibit longer tails, suggesting that each silo's performance is influenced by the specific data it has access to.
\begin{table}[]
\centering
\small
\caption{The mean and delta of the F1 value using standard configurations. By examining the mean column, it is evident that our CFL model yields a higher F1 score. In the delta column, our classifier shows a smaller difference, indicating that its F1 score is closely aligned with that of the global (imaginary) data. A positive value indicates that the F1 score outperforms that of the global (imaginary) data.}
\begin{tabular}{|l|rrrrr|rrrr|}
\hline
\multicolumn{1}{|c|}{Dataset} & \multicolumn{5}{c|}{Mean}                                                                                                                                                                                                                      & \multicolumn{4}{c|}{Delta}                                                                                                                                                                                                                                                                                                              \\ \hline
                              & \multicolumn{1}{c|}{Base1} & \multicolumn{1}{c|}{\begin{tabular}[c]{@{}c@{}}CFL \\ (Ours)\end{tabular}} & \multicolumn{1}{c|}{Base2} & \multicolumn{1}{c|}{SubTab} & \multicolumn{1}{c|}{\begin{tabular}[c]{@{}c@{}}SubTab \\ FL\end{tabular}} & \multicolumn{1}{c|}{\begin{tabular}[c]{@{}c@{}}CFL\\ -\\ Base1\end{tabular}} & \multicolumn{1}{c|}{\begin{tabular}[c]{@{}c@{}}Base2\\ -\\ Base1\end{tabular}} & \multicolumn{1}{c|}{\begin{tabular}[c]{@{}c@{}}SubTab\\ -\\ Base1\end{tabular}} & \multicolumn{1}{c|}{\begin{tabular}[c]{@{}c@{}}SubTab\\ FL\\ -\\ Base 1\end{tabular}} \\ \hline
blog                                           & \multicolumn{1}{r|}{0.791}  & \multicolumn{1}{r|}{0.774}                                                 & \multicolumn{1}{r|}{0.760}  & \multicolumn{1}{r|}{0.757}  & 0.724                                                                     & \multicolumn{1}{r|}{-0.017}                                               & \multicolumn{1}{r|}{-0.031}                                              & \multicolumn{1}{r|}{-0.035}                                                  & -0.068                                                                         \\ \hline
covtype                                        & \multicolumn{1}{r|}{0.561}  & \multicolumn{1}{r|}{0.527}                                                 & \multicolumn{1}{r|}{0.367}  & \multicolumn{1}{r|}{0.357}  & 0.346                                                                     & \multicolumn{1}{r|}{-0.033}                                               & \multicolumn{1}{r|}{-0.193}                                              & \multicolumn{1}{r|}{-0.204}                                                  & -0.214                                                                         \\ \hline
income                                         & \multicolumn{1}{r|}{0.811}  & \multicolumn{1}{r|}{0.739}                                                 & \multicolumn{1}{r|}{0.674}  & \multicolumn{1}{r|}{0.726}  & 0.662                                                                     & \multicolumn{1}{r|}{-0.072}                                               & \multicolumn{1}{r|}{-0.136}                                              & \multicolumn{1}{r|}{-0.085}                                                  & -0.149                                                                         \\ \hline
sensorless                                     & \multicolumn{1}{r|}{0.154}  & \multicolumn{1}{r|}{0.565}                                                 & \multicolumn{1}{r|}{0.075}  & \multicolumn{1}{r|}{0.080}  & 0.067                                                                     & \multicolumn{1}{r|}{0.411}                                                & \multicolumn{1}{r|}{-0.079}                                              & \multicolumn{1}{r|}{-0.075}                                                  & -0.088                                                                         \\ \hline
syn                                            & \multicolumn{1}{r|}{0.859}  & \multicolumn{1}{r|}{0.773}                                                 & \multicolumn{1}{r|}{0.776}  & \multicolumn{1}{r|}{0.403}  & 0.368                                                                     & \multicolumn{1}{r|}{-0.086}                                               & \multicolumn{1}{r|}{-0.083}                                              & \multicolumn{1}{r|}{-0.456}                                                  & -0.491                                                                         \\ \hline
tuandromd                                      & \multicolumn{1}{r|}{0.939}  & \multicolumn{1}{r|}{0.833}                                                 & \multicolumn{1}{r|}{0.701}  & \multicolumn{1}{r|}{0.704}  & 0.703                                                                     & \multicolumn{1}{r|}{-0.105}                                               & \multicolumn{1}{r|}{-0.237}                                              & \multicolumn{1}{r|}{-0.235}                                                  & -0.235                                                                         \\ \hline
\end{tabular}
\label{tab:subtabResume}
\end{table}

\subsection{Data Size Imbalance Setting}
\begin{figure}[]
    \centering
    \begin{minipage}{0.47\linewidth}
        \centering
        \includegraphics[width=1\linewidth]{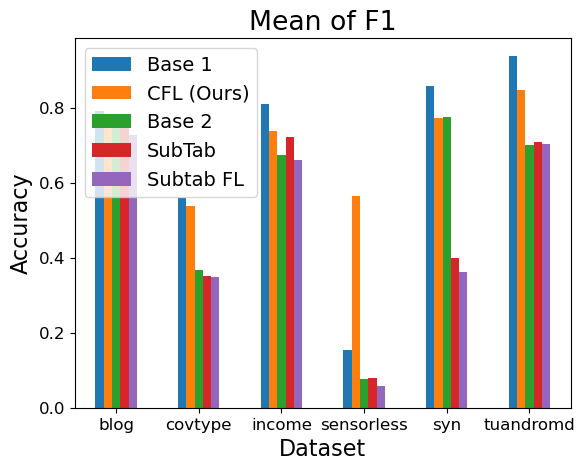} 
        \caption{Bar graphs depicting the average of F1 scores under 'Data Size Imbalance Setting'. In most instances, CFL outperforms other models. However, there is an exception in the syn dataset where CFL does not enhance the performance of the Base 2 (local) model.}
        \label{img:dataDropResume}
    \end{minipage}\hfill
    \begin{minipage}{0.47\linewidth}
        \centering
        \includegraphics[width=1\linewidth]{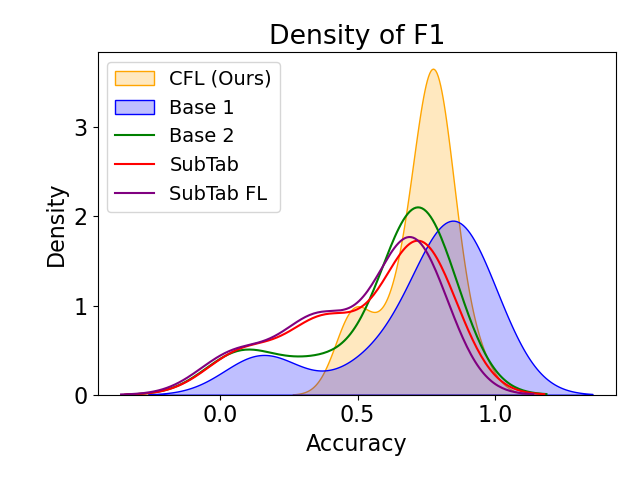} 
        \caption{Our CFL demonstrates a higher Kernel Density Estimate (KDE) of F1 scores in comparison to other models, with a reduced tail length. This indicates that the CFL results in less variability.}
        \label{img:dataDropDensity}
    \end{minipage}
\end{figure}

The results of the experiments conducted are similar to those of the standard setting described earlier. Our CFL model shows superior performance across all datasets except for the syn dataset. In addition, compared to the local model Base 2, the increase in recall is more significant than the increase in precision. The SubTab model shows improvements in performance in the sensorless, income, and tuandromd datasets compared to the local model Base 2. However, our CFL model outperforms both. The detailed results of the experiments are presented in Table \ref{table:table_client_drop}. The mean F1 scores, as shown in Table \ref{table:table_client_dropResume} and Figure \ref{img:dataDropResume}, demonstrate the improved performance of both CFL and SubTab. SubTab, which adopts FL (SubTab-FL), performs below the local Base 2 model. Results similar to those of the standard setting are shown in Figure \ref{img:dataDropDensity}.

\begin{table}[]
\caption{The mean and delta of the F1 value using 'Data Size Imbalance Setting'. Upon examination of the mean column, it is evident that our CFL model yields a higher F1 score. In the delta column, our CFL shows a lower value, indicating that its F1 score is closely aligned with that of the global (imaginary) data. A positive value indicates that the F1 score outperforms that of the global (imaginary) data.}
\centering
\small
\begin{tabular}{|l|rrrrr|rrrr|}
\hline
\multicolumn{1}{|c|}{Dataset} & \multicolumn{5}{c|}{Mean}                                                                                                                                                                                                                      & \multicolumn{4}{c|}{Delta}                                                                                                                                                                                                                                                                                                              \\ \hline
                              & \multicolumn{1}{c|}{Base1} & \multicolumn{1}{c|}{\begin{tabular}[c]{@{}c@{}}CFL \\ (Ours)\end{tabular}} & \multicolumn{1}{c|}{Base2} & \multicolumn{1}{c|}{SubTab} & \multicolumn{1}{c|}{\begin{tabular}[c]{@{}c@{}}SubTab \\ FL\end{tabular}} & \multicolumn{1}{c|}{\begin{tabular}[c]{@{}c@{}}CFL\\ -\\ Base1\end{tabular}} & \multicolumn{1}{c|}{\begin{tabular}[c]{@{}c@{}}Base2\\ -\\ Base1\end{tabular}} & \multicolumn{1}{c|}{\begin{tabular}[c]{@{}c@{}}SubTab\\ -\\ Base1\end{tabular}} & \multicolumn{1}{c|}{\begin{tabular}[c]{@{}c@{}}SubTab\\ FL\\ -\\ Base 1\end{tabular}} \\ \hline
blog                                           & \multicolumn{1}{r|}{0.791}  & \multicolumn{1}{r|}{0.776}                                                 & \multicolumn{1}{r|}{0.760}  & \multicolumn{1}{r|}{0.757}  & 0.727                                                                     & \multicolumn{1}{r|}{-0.016}                                                   & \multicolumn{1}{r|}{-0.031}                                                      & \multicolumn{1}{r|}{-0.035}                                                      & -0.064                                                                             \\ \hline
covtype                                        & \multicolumn{1}{r|}{0.561}  & \multicolumn{1}{r|}{0.539}                                                 & \multicolumn{1}{r|}{0.367}  & \multicolumn{1}{r|}{0.352}  & 0.349                                                                     & \multicolumn{1}{r|}{-0.022}                                                   & \multicolumn{1}{r|}{-0.193}                                                      & \multicolumn{1}{r|}{-0.209}                                                      & -0.211                                                                             \\ \hline
income                                         & \multicolumn{1}{r|}{0.811}  & \multicolumn{1}{r|}{0.739}                                                 & \multicolumn{1}{r|}{0.674}  & \multicolumn{1}{r|}{0.722}  & 0.662                                                                     & \multicolumn{1}{r|}{-0.071}                                                   & \multicolumn{1}{r|}{-0.136}                                                      & \multicolumn{1}{r|}{-0.088}                                                      & -0.149                                                                             \\ \hline
sensorless                                     & \multicolumn{1}{r|}{0.154}  & \multicolumn{1}{r|}{0.564}                                                 & \multicolumn{1}{r|}{0.075}  & \multicolumn{1}{r|}{0.080}  & 0.058                                                                     & \multicolumn{1}{r|}{0.409}                                                    & \multicolumn{1}{r|}{-0.079}                                                      & \multicolumn{1}{r|}{-0.075}                                                      & -0.097                                                                             \\ \hline
syn                                            & \multicolumn{1}{r|}{0.859}  & \multicolumn{1}{r|}{0.773}                                                 & \multicolumn{1}{r|}{0.776}  & \multicolumn{1}{r|}{0.400}  & 0.362                                                                     & \multicolumn{1}{r|}{-0.086}                                                   & \multicolumn{1}{r|}{-0.083}                                                      & \multicolumn{1}{r|}{-0.459}                                                      & -0.497                                                                             \\ \hline
tuandromd                                      & \multicolumn{1}{r|}{0.939}  & \multicolumn{1}{r|}{0.848}                                                 & \multicolumn{1}{r|}{0.701}  & \multicolumn{1}{r|}{0.710}  & 0.703                                                                     & \multicolumn{1}{r|}{-0.090}                                                   & \multicolumn{1}{r|}{-0.237}                                                      & \multicolumn{1}{r|}{-0.229}                                                      & -0.235                                                                             \\ \hline
\end{tabular}
\label{table:table_client_dropResume}
\end{table}

\subsection{Class Size Imbalance Setting}
\begin{figure}[]
    \centering
    \begin{minipage}{0.47\linewidth}
        \centering
        \includegraphics[width=1\linewidth]{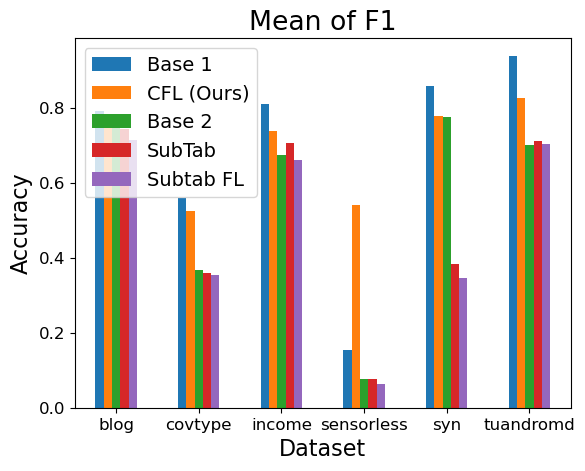} 
        \caption{Bar graphs depicting the average of F1 scores under 'Class Size Imbalance Setting'. In all instances, CFL outperforms other models. }
        \label{img:classImbalance}
    \end{minipage}\hfill
    \begin{minipage}{0.47\linewidth}
        \centering
        \includegraphics[width=1\linewidth]{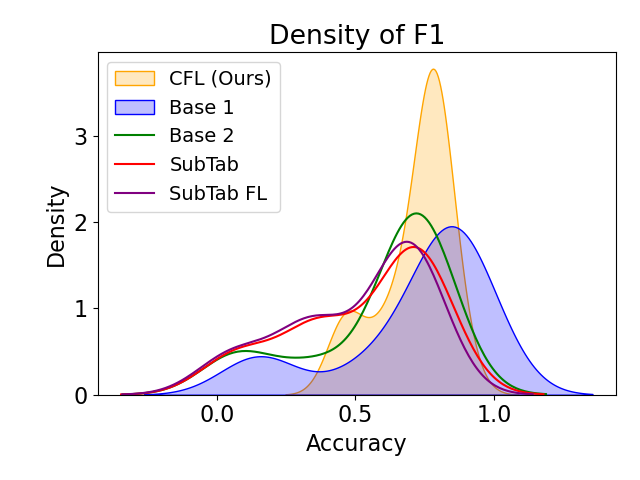} 
        \caption{Our CFL exhibits a higher Kernel Density Estimate (KDE) of F1 scores in comparison to other models, with a reduced tail length. This indicates that the CFL results in less variability.}
        \label{img:classDensity}
    \end{minipage}
\end{figure}

The performance of our CFL closely resembles that of the global model Base 1 when applied across all six datasets. However, unlike in the previous scenarios, our CFL improves performance on the syn dataset. The results of these experiments are detailed in Table \ref{table:class_imbalance}. Table \ref{table:class_imbalanceResume} and Figure \ref{img:classImbalance} show the mean F1 scores of each model. Our CFL demonstrates the closest performance to outperforming the global model Base 1. Figure \ref{img:classDensity} presents results that are consistent with those of previous experiments.

\begin{table}[]
\caption{The mean and delta of the F1 value using 'Class Size Imbalance Setting'. By examining the mean column, it is evident that our CFL model yields a higher F1 score. In the delta column, our classifier shows a smaller difference, indicating that its F1 score is closely aligned with that of the global (imaginary) data. A positive value indicates that the F1 score outperforms that of the global (imaginary) data.}
\centering
\small
\begin{tabular}{|l|rrrrr|rrrr|}
\hline
\multicolumn{1}{|c|}{Dataset} & \multicolumn{5}{c|}{Mean}                                                                                                                                                                                                                      & \multicolumn{4}{c|}{Delta}                                                                                                                                                                                                                                                                                                              \\ \hline
                              & \multicolumn{1}{c|}{Base1} & \multicolumn{1}{c|}{\begin{tabular}[c]{@{}c@{}}CFL \\ (Ours)\end{tabular}} & \multicolumn{1}{c|}{Base2} & \multicolumn{1}{c|}{SubTab} & \multicolumn{1}{c|}{\begin{tabular}[c]{@{}c@{}}SubTab \\ FL\end{tabular}} & \multicolumn{1}{c|}{\begin{tabular}[c]{@{}c@{}}CFL\\ -\\ Base1\end{tabular}} & \multicolumn{1}{c|}{\begin{tabular}[c]{@{}c@{}}Base2\\ -\\ Base1\end{tabular}} & \multicolumn{1}{c|}{\begin{tabular}[c]{@{}c@{}}SubTab\\ -\\ Base1\end{tabular}} & \multicolumn{1}{c|}{\begin{tabular}[c]{@{}c@{}}SubTab\\ FL\\ -\\ Base 1\end{tabular}} \\ \hline
blog                                           & \multicolumn{1}{r|}{0.791}  & \multicolumn{1}{r|}{0.775}                                                 & \multicolumn{1}{r|}{0.760}  & \multicolumn{1}{r|}{0.744}  & 0.713                                                                     & \multicolumn{1}{r|}{-0.017}                                                   & \multicolumn{1}{r|}{-0.031}                                                      & \multicolumn{1}{r|}{-0.048}                                                      & -0.078                                                                             \\ \hline
covtype                                        & \multicolumn{1}{r|}{0.561}  & \multicolumn{1}{r|}{0.526}                                                 & \multicolumn{1}{r|}{0.367}  & \multicolumn{1}{r|}{0.358}  & 0.355                                                                     & \multicolumn{1}{r|}{-0.035}                                                   & \multicolumn{1}{r|}{-0.193}                                                      & \multicolumn{1}{r|}{-0.202}                                                      & -0.205                                                                             \\ \hline
income                                         & \multicolumn{1}{r|}{0.811}  & \multicolumn{1}{r|}{0.739}                                                 & \multicolumn{1}{r|}{0.674}  & \multicolumn{1}{r|}{0.707}  & 0.662                                                                     & \multicolumn{1}{r|}{-0.071}                                                   & \multicolumn{1}{r|}{-0.136}                                                      & \multicolumn{1}{r|}{-0.103}                                                      & -0.149                                                                             \\ \hline
sensorless                                     & \multicolumn{1}{r|}{0.154}  & \multicolumn{1}{r|}{0.539}                                                 & \multicolumn{1}{r|}{0.075}  & \multicolumn{1}{r|}{0.076}  & 0.064                                                                     & \multicolumn{1}{r|}{0.385}                                                    & \multicolumn{1}{r|}{-0.079}                                                      & \multicolumn{1}{r|}{-0.078}                                                      & -0.090                                                                             \\ \hline
syn                                            & \multicolumn{1}{r|}{0.859}  & \multicolumn{1}{r|}{0.779}                                                 & \multicolumn{1}{r|}{0.776}  & \multicolumn{1}{r|}{0.383}  & 0.346                                                                     & \multicolumn{1}{r|}{-0.080}                                                   & \multicolumn{1}{r|}{-0.083}                                                      & \multicolumn{1}{r|}{-0.477}                                                      & -0.513                                                                             \\ \hline
tuandromd                                      & \multicolumn{1}{r|}{0.939}  & \multicolumn{1}{r|}{0.826}                                                 & \multicolumn{1}{r|}{0.701}  & \multicolumn{1}{r|}{0.711}  & 0.703                                                                     & \multicolumn{1}{r|}{-0.112}                                                   & \multicolumn{1}{r|}{-0.237}                                                      & \multicolumn{1}{r|}{-0.228}                                                      & -0.235                                                                             \\ \hline
\end{tabular}
\label{table:class_imbalanceResume}
\end{table}

\subsection{Mixed Case Settings}
Our CFL has proven to be highly effective in this setting. The detailed results of the experiments are presented in Table \ref{tab:mixSetup}. Our CFL outperforms the Local Base 2 model on all six datasets, performing significantly better and closely resembling the Global Base 1 model. However, the Local SubTab and SubTab FL do not show a significant improvement over the Local Base 2 model.

\begin{figure}[]
    \centering
    \begin{minipage}{0.47\linewidth}
        \centering
        \includegraphics[width=1\linewidth]{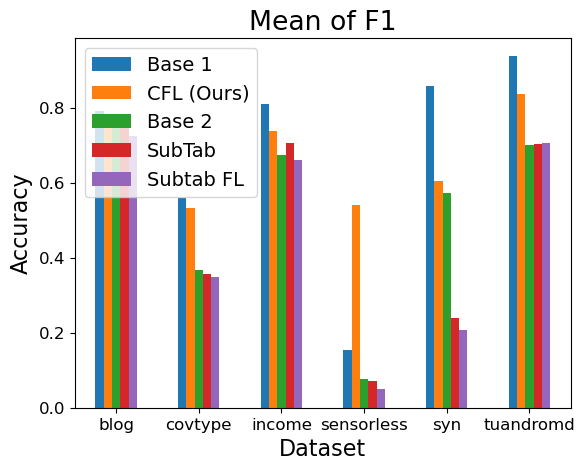} 
        \caption{Bar graphs depicting the average of F1 scores under 'Mixed Case Settings'. In all instances, the CFL model surpasses the other models. In contrast to other scenarios, the CFL model outperforms other models by a greater margin, even within the syn dataset.}
        \label{img:mixSetting}
    \end{minipage}\hfill
    \begin{minipage}{0.47\linewidth}
        \centering
        \includegraphics[width=1\linewidth]{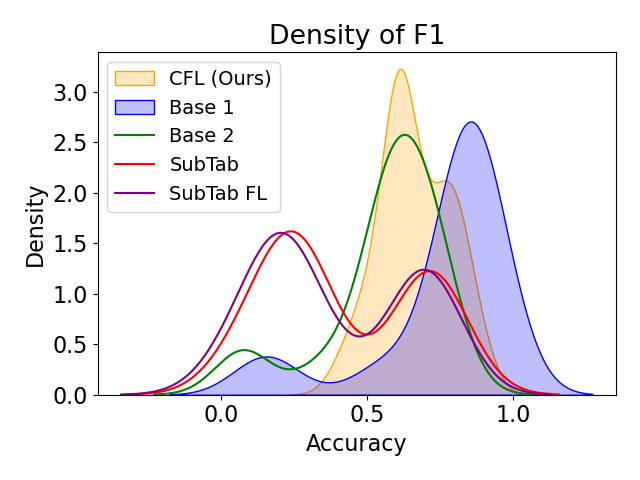} 
        \caption{Our CFL shows a higher Figure of Density (KDE) of F1 scores compared to other models, with a shorter tail. This means that CFL gives better F1 scores with smaller variations.}
        \label{img:mixSettingDensity}
    \end{minipage}
\end{figure}

You can see this comparison in the Table \ref{tab:mixedSettingResume} and Figure \ref{img:mixSetting}. The density plot in Figure \ref{img:mixSettingDensity} shows distinct outcomes compared to the previous experiment. Despite having a higher density, the CFL performs similarly to the other models in terms of density because of variations in settings across different contexts. However, the CFL maintains a shorter tail, indicating less variability in F1 scores. 

\begin{table}[]
\caption{The mean and delta of the F1 value using 'Mixed Case Settings'. Upon examination of the mean column, it is evident that our CFL model yields a higher F1 score. In the delta column, our classifier shows a smaller difference, indicating that its F1 score is closely aligned with that of the global (imaginary) data. A positive value indicates that the F1 score outperforms that of the global (imaginary) data.}
\centering
\small
\begin{tabular}{|l|rrrrr|rrrr|}
\hline
\multicolumn{1}{|c|}{Dataset} & \multicolumn{5}{c|}{Mean}                                                                                                                                                                                                                      & \multicolumn{4}{c|}{Delta}                                                                                                                                                                                                                                                                                                              \\ \hline
                              & \multicolumn{1}{c|}{Base1} & \multicolumn{1}{c|}{\begin{tabular}[c]{@{}c@{}}CFL \\ (Ours)\end{tabular}} & \multicolumn{1}{c|}{Base2} & \multicolumn{1}{c|}{SubTab} & \multicolumn{1}{c|}{\begin{tabular}[c]{@{}c@{}}SubTab \\ FL\end{tabular}} & \multicolumn{1}{c|}{\begin{tabular}[c]{@{}c@{}}CFL\\ -\\ Base1\end{tabular}} & \multicolumn{1}{c|}{\begin{tabular}[c]{@{}c@{}}Base2\\ -\\ Base1\end{tabular}} & \multicolumn{1}{c|}{\begin{tabular}[c]{@{}c@{}}SubTab\\ -\\ Base1\end{tabular}} & \multicolumn{1}{c|}{\begin{tabular}[c]{@{}c@{}}SubTab\\ FL\\ -\\ Base 1\end{tabular}} \\ \hline
blog                                           & \multicolumn{1}{r|}{0.791}  & \multicolumn{1}{r|}{0.773}                                                 & \multicolumn{1}{r|}{0.760}  & \multicolumn{1}{r|}{0.749}  & 0.723                                                                     & \multicolumn{1}{r|}{-0.018}                                                   & \multicolumn{1}{r|}{-0.031}                                                      & \multicolumn{1}{r|}{-0.042}                                                      & -0.068                                                                             \\ \hline
covtype                                        & \multicolumn{1}{r|}{0.561}  & \multicolumn{1}{r|}{0.533}                                                 & \multicolumn{1}{r|}{0.367}  & \multicolumn{1}{r|}{0.358}  & 0.348                                                                     & \multicolumn{1}{r|}{-0.027}                                                   & \multicolumn{1}{r|}{-0.193}                                                      & \multicolumn{1}{r|}{-0.203}                                                      & -0.212                                                                             \\ \hline
income                                         & \multicolumn{1}{r|}{0.811}  & \multicolumn{1}{r|}{0.739}                                                 & \multicolumn{1}{r|}{0.674}  & \multicolumn{1}{r|}{0.705}  & 0.662                                                                     & \multicolumn{1}{r|}{-0.071}                                                   & \multicolumn{1}{r|}{-0.136}                                                      & \multicolumn{1}{r|}{-0.106}                                                      & -0.149                                                                             \\ \hline
sensorless                                     & \multicolumn{1}{r|}{0.154}  & \multicolumn{1}{r|}{0.540}                                                 & \multicolumn{1}{r|}{0.075}  & \multicolumn{1}{r|}{0.070}  & 0.050                                                                     & \multicolumn{1}{r|}{0.386}                                                    & \multicolumn{1}{r|}{-0.079}                                                      & \multicolumn{1}{r|}{-0.084}                                                      & -0.104                                                                             \\ \hline
syn                                            & \multicolumn{1}{r|}{0.859}  & \multicolumn{1}{r|}{0.604}                                                 & \multicolumn{1}{r|}{0.573}  & \multicolumn{1}{r|}{0.240}  & 0.208                                                                     & \multicolumn{1}{r|}{-0.256}                                                   & \multicolumn{1}{r|}{-0.287}                                                      & \multicolumn{1}{r|}{-0.619}                                                      & -0.651                                                                             \\ \hline
tuandromd                                      & \multicolumn{1}{r|}{0.939}  & \multicolumn{1}{r|}{0.836}                                                 & \multicolumn{1}{r|}{0.701}  & \multicolumn{1}{r|}{0.703}  & 0.706                                                                     & \multicolumn{1}{r|}{-0.103}                                                   & \multicolumn{1}{r|}{-0.237}                                                      & \multicolumn{1}{r|}{-0.235}                                                      & -0.233                                                                             \\ \hline
\end{tabular}
\label{tab:mixedSettingResume}
\end{table}

\subsection{Effects of the Pearson Reordering}
Our CFL introduced a simple Pearson reordering technique. Table \ref{tab:pearsonAccuracy} shows the experiment to test the effectiveness of our proposed model.  The experiment was done under the standard setting. Pearson reordering improved performance across all datasets, as indicated by the mean performance scores presented in the table. All models showed improvements in precision, recall and F1 scores, with significant enhancements observed in the income, covtype, and sensorless datasets. The impact was particularly notable in the sensorless dataset. Our experiments demonstrated consistent effectiveness in various datasets.
\begin{table}[]
\centering
\small
\caption{Mean of performance of our CFL when Pearson Reordering is applied and not applied under standard setting.}
\begin{tabular}{|l|rrr|rrr|}
\hline
\multicolumn{1}{|c|}{Dataset} & \multicolumn{3}{c|}{With Pearson Reordering}                                             & \multicolumn{3}{c|}{Without Pearson Reordering}                                          \\ \hline
                              & \multicolumn{1}{c|}{Precision} & \multicolumn{1}{c|}{Recal}    & \multicolumn{1}{c|}{F1} & \multicolumn{1}{c|}{Precision} & \multicolumn{1}{c|}{Recal}    & \multicolumn{1}{c|}{F1} \\ \hline
blog                                           & \multicolumn{1}{r|}{0.774908}  & \multicolumn{1}{r|}{0.776912} & 0.774445                & \multicolumn{1}{r|}{0.763213}  & \multicolumn{1}{r|}{0.763242} & 0.762656                \\ \hline
covtype                                        & \multicolumn{1}{r|}{0.556630}  & \multicolumn{1}{r|}{0.576441} & 0.531228                & \multicolumn{1}{r|}{0.396080}  & \multicolumn{1}{r|}{0.509270} & 0.386813                \\ \hline
income                                         & \multicolumn{1}{r|}{0.767016}  & \multicolumn{1}{r|}{0.783577} & 0.739005                & \multicolumn{1}{r|}{0.726834}  & \multicolumn{1}{r|}{0.774993} & 0.709457                \\ \hline
sensorless                                     & \multicolumn{1}{r|}{0.562038}  & \multicolumn{1}{r|}{0.562557} & 0.552093                & \multicolumn{1}{r|}{0.114364}  & \multicolumn{1}{r|}{0.129997} & 0.083378                \\ \hline
syn                                            & \multicolumn{1}{r|}{0.690266}  & \multicolumn{1}{r|}{0.685348} & 0.677071                & \multicolumn{1}{r|}{0.687704}  & \multicolumn{1}{r|}{0.685320} & 0.677161                \\ \hline
tuandromd                                      & \multicolumn{1}{r|}{0.866893}  & \multicolumn{1}{r|}{0.861587} & 0.836021                & \multicolumn{1}{r|}{0.777319}  & \multicolumn{1}{r|}{0.795953} & 0.708342                \\ \hline
\end{tabular}
\label{tab:pearsonAccuracy}
\end{table}
\subsection{Effects of dot product loss.}
In order to accelerate the model performance, our CFL incorporates a dot product for contrastive loss. Table \ref{tab:cdot} compares the performance of the dot product and cosine within the loss function under normal settings. The results show the time in seconds that CFL takes to process one batch. The table clearly demonstrates that the dot product consistently yields shorter processing times across all tested datasets. By utilizing the dot product, our CFL has successfully reduced training time significantly.
\begin{table}[]
    \centering
    \caption{Comparison of the implementation time for dot product and cos. The time provided (in seconds) is the average time for a single epoch. The ratio is calculated based on $\frac{t_{cos}}{t_{dot}}$ where $t$ is time. From the table, we can see that the implementation of the dot product adds speed to our CFL.}
    \begin{tabular}{|l|rrrrrr|} \hline
         &  Blog&  Covtype&  Income&  Sensorless&  Synthetic & Tuandromd\\ \hline
         Dot product&  18&  1.5e3&  6&  18&  41& 1\\
 Cosine& 74& 1.9e3& 40& 101& 93&5\\ \hline
 Ratio& 4.1& 1.2& 6.7& 5.6& 2.3&5\\ \hline
    \end{tabular}
    
    \label{tab:cdot}
\end{table}
\subsection{Comparison with  Deep Learning Algorithms}

Results on Table \ref{tab:gbdt_results} demonstrate that CFLc consistently outperforms other deep learning approaches (Scarf, MLP, and Transformer) across all datasets, with particularly significant performance gaps in the sensorless and synthetic datasets. Moreover, when CFL is enhanced with LightGBM (CFL+$^z$), it shows even more impressive improvements, especially in datasets like sensorless (0.565 to 0.812), covtype (0.527 to 0.650), and tuandromd (0.833 to 0.932). The Transformer model notably struggles with certain datasets, achieving very low scores on sensorless (0.014) and synthetic (0.060) datasets. This comparison effectively demonstrates CFL's capability in handling federated learning scenarios compared to other deep learning approaches while also showing that its performance can be further enhanced through integration with other techniques like LightGBM.
\begin{table}
    \centering
    \caption{Performance comparison between CFL and other deep learning models. $^c$ is for contrastive learning and $^z$ is for CFL+LightGBM.}
    \begin{tabular}{|l|cccccc|} \hline 
         Method&  blog&  covtype&  income&  sensorless&  syn& tuandromd\\ \hline
         CFL$^c$&  \textbf{0.774}&  \textbf{0.527} &  \textbf{0.739} &  \textbf{0.565} &  \textbf{0.773}& \textbf{0.833} \\
 Scarf& 0.675& 0.386& 0.691& 0.061& 0.263&0.762\\
 MLP& 0.740& 0.510& 0.736& 0.495& 0.685&0.701\\
 Transformer& 0.765& 0.319& 0.710& 0.014& 0.060&0.641\\ \hline
 CFL+$^{z}$& 0.771& \textbf{0.650} & \textbf{0.750}& \textbf{0.812} & 0.651&\textbf{0.932}\\ \hline 
    \end{tabular}
     
    \label{tab:gbdt_results}
\end{table}
\subsection{ Evaluation}
The CFL algorithm has been shown to perform better than local models and, in some cases, even better than models trained on global datasets. Our CFL algorithm performs well in situations where there is both data size imbalance and class imbalance in the datasets. This improvement is evident in increased precision and recall. Although the improvements in precision are small, the gains in recall are significant. As a result, both metrics show improvement, indicating that CFL significantly enhances the model's ability to detect positive instances while reducing false positive predictions. 

\begin{figure}[H]
    \centering
    \includegraphics[width=0.75\linewidth, height=5cm]{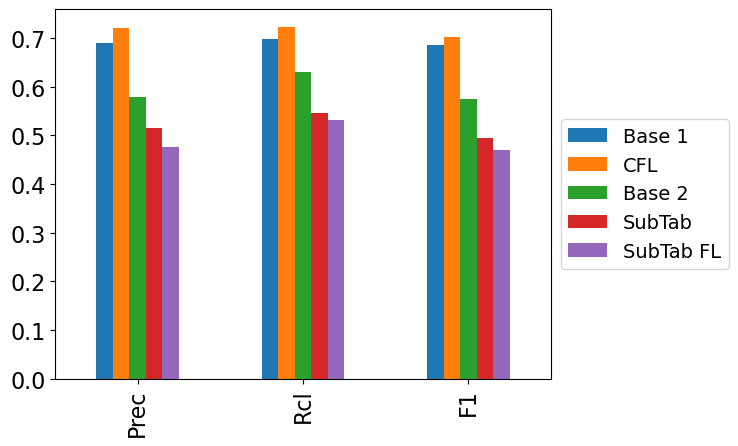}
    \caption{The mean of the matrix for each model indicates that our CFL model outperforms all other models, including the one trained with a global (imaginary) dataset.}
    \label{fig:delta}
\end{figure}

Our full matrix representation, simple Pearson reordering, and dot product loss calculation approach have been instrumental in achieving the outcomes. While the dot product improves training speed, person reordering improves performance. The mean values of the matrix of each model are shown in Figure \ref{fig:delta}, which illustrates that, in general, the CFL outperforms all models, including those trained on global (imaginary) datasets.
\section{Conclusion}
Challenges such as vertical partition, sample misalignment, and privacy constraints are often real-world scenarios that define the importance of our study. Our CFL provides an approach to learning tabular data from data silos, and we are the first to offer a simultaneous approach to the above problems. Our CFL is capable of adapting and maintaining higher performance even in complex situations. By maintaining privacy constraints that are typically considered in federated learning research, CFL opens new possibilities for collaborative learning in highly sensitive environments.

Although CFL shows promising results, there are limitations and areas for further research. 
Future work could focus on optimizing CFL for larger-scale deployments with more silos and larger datasets. Exploring CFL's adaptability to changing data distributions and evolving silo structures could enhance its real-world applicability. Developing methods to improve the interpretability of CFL's decision-making process could increase trust and adoption in sensitive sectors. Investigating CFL's potential in other domains beyond tabular data, such as text or time series data, could broaden its impact.

\section{Acknowledgements}
This work acknowledges the funding provided by the Indonesia Endowment Funds for Education, as well as the support from the Indonesian Taxation Office and the University of Queensland, Australia.

\bibliographystyle{elsarticle-harv}
\bibliography{export}

\appendix
\section{Experiment Results \label{apdx}}
\subsection{Standard Setting \label{appendix:rawStandard}}

\begin{sidewaystable}[htbp]
    \caption{\textbf{A.1} Table of precision, recall, and F1 of each model on experimented datasets within standard settings. Our CFL able to maintain higher performance. Compared to other models, our CFL gains a significant increase in recall. The precision also improved.} \scriptsize
\begin{tabular}{|l|l|lll|lll|lll|lll|lll|}
\hline
Model      & Model  & \multicolumn{3}{l|}{Base 1} & \multicolumn{3}{l|}{CFL (Ours)} & \multicolumn{3}{l|}{Base 2} & \multicolumn{3}{l|}{SubTab} & \multicolumn{3}{l|}{SubTab-FL} \\ \hline
dataset    & client & Prec    & Recal    & F1     & Prec     & Recal     & F1       & Prec    & Recal    & F1     & Prec    & Recal    & F1     & Prec     & Recal     & F1      \\ \hline
blog       & 1      & 0.79    & 0.80     & 0.79   & 0.74     & 0.76      & 0.74     & 0.72    & 0.72     & 0.72   & 0.72    & 0.73     & 0.72   & 0.71     & 0.68      & 0.69    \\
blog       & 2      & 0.79    & 0.80     & 0.79   & 0.79     & 0.78      & 0.79     & 0.77    & 0.77     & 0.77   & 0.77    & 0.77     & 0.77   & 0.73     & 0.71      & 0.72    \\
blog       & 3      & 0.79    & 0.80     & 0.79   & 0.81     & 0.80      & 0.80     & 0.80    & 0.79     & 0.79   & 0.79    & 0.78     & 0.78   & 0.77     & 0.75      & 0.76    \\
blog       & 4      & 0.79    & 0.80     & 0.79   & 0.76     & 0.76      & 0.76     & 0.75    & 0.76     & 0.76   & 0.75    & 0.74     & 0.75   & 0.74     & 0.72      & 0.73    \\ \hline
covtype    & 1      & 0.55    & 0.60     & 0.56   & 0.49     & 0.54      & 0.47     & 0.24    & 0.49     & 0.32   & 0.38    & 0.46     & 0.35   & 0.23     & 0.48      & 0.31    \\
covtype    & 2      & 0.55    & 0.60     & 0.56   & 0.56     & 0.56      & 0.50     & 0.48    & 0.49     & 0.36   & 0.51    & 0.49     & 0.34   & 0.47     & 0.49      & 0.34    \\
covtype    & 3      & 0.55    & 0.60     & 0.56   & 0.61     & 0.63      & 0.61     & 0.47    & 0.51     & 0.42   & 0.42    & 0.49     & 0.38   & 0.41     & 0.49      & 0.39    \\ \hline
income     & 1      & 0.82    & 0.83     & 0.81   & 0.77     & 0.79      & 0.75     & 0.56    & 0.75     & 0.64   & 0.77    & 0.78     & 0.75   & 0.81     & 0.75      & 0.64    \\
income     & 2      & 0.82    & 0.83     & 0.81   & 0.74     & 0.76      & 0.68     & 0.66    & 0.75     & 0.65   & 0.72    & 0.76     & 0.68   & 0.56     & 0.75      & 0.64    \\
income     & 3      & 0.82    & 0.83     & 0.81   & 0.73     & 0.76      & 0.71     & 0.56    & 0.75     & 0.64   & 0.72    & 0.75     & 0.67   & 0.56     & 0.75      & 0.64    \\
income     & 4      & 0.82    & 0.83     & 0.81   & 0.80     & 0.80      & 0.76     & 0.78    & 0.77     & 0.71   & 0.81    & 0.79     & 0.75   & 0.76     & 0.77      & 0.69    \\
income     & 5      & 0.82    & 0.83     & 0.81   & 0.80     & 0.81      & 0.79     & 0.78    & 0.78     & 0.73   & 0.80    & 0.81     & 0.79   & 0.78     & 0.77      & 0.69    \\ \hline
sensorless & 1      & 0.17    & 0.16     & 0.15   & 0.52     & 0.50      & 0.49     & 0.27    & 0.10     & 0.05   & 0.10    & 0.15     & 0.10   & 0.07     & 0.11      & 0.05    \\
sensorless & 2      & 0.17    & 0.16     & 0.15   & 0.71     & 0.70      & 0.69     & 0.11    & 0.12     & 0.07   & 0.13    & 0.13     & 0.11   & 0.16     & 0.15      & 0.12    \\
sensorless & 3      & 0.17    & 0.16     & 0.15   & 0.61     & 0.62      & 0.61     & 0.16    & 0.13     & 0.10   & 0.06    & 0.10     & 0.05   & 0.06     & 0.10      & 0.04    \\
sensorless & 4      & 0.17    & 0.16     & 0.15   & 0.47     & 0.48      & 0.47     & 0.14    & 0.12     & 0.08   & 0.08    & 0.11     & 0.07   & 0.06     & 0.10      & 0.10    \\ \hline
syn        & 1      & 0.86    & 0.86     & 0.86   & 0.78     & 0.78      & 0.77     & 0.78    & 0.78     & 0.78   & 0.40    & 0.40     & 0.39   & 0.39     & 0.39      & 0.37    \\
syn        & 2      & 0.86    & 0.86     & 0.86   & 0.78     & 0.77      & 0.76     & 0.77    & 0.77     & 0.77   & 0.42    & 0.42     & 0.42   & 0.39     & 0.39      & 0.38    \\
syn        & 3      & 0.86    & 0.86     & 0.86   & 0.79     & 0.79      & 0.78     & 0.79    & 0.79     & 0.78   & 0.42    & 0.42     & 0.41   & 0.39     & 0.38      & 0.37    \\
syn        & 4      & 0.86    & 0.86     & 0.86   & 0.79     & 0.78      & 0.77     & 0.78    & 0.78     & 0.77   & 0.39    & 0.40     & 0.39   & 0.36     & 0.36      & 0.35    \\ \hline
tuandromd  & 1      & 0.94    & 0.94     & 0.94   & 0.87     & 0.87      & 0.85     & 0.63    & 0.79     & 0.70   & 0.84    & 0.79     & 0.70   & 0.63     & 0.79      & 0.70    \\ 
tuandromd  & 2      & 0.94    & 0.94     & 0.94   & 0.84     & 0.85      & 0.83     & 0.63    & 0.79     & 0.70   & 0.63    & 0.79     & 0.70   & 0.63     & 0.79      & 0.70    \\
tuandromd  & 3      & 0.94    & 0.94     & 0.94   & 0.90     & 0.90      & 0.89     & 0.63    & 0.79     & 0.70   & 0.63    & 0.79     & 0.70   & 0.63     & 0.79      & 0.70    \\
tuandromd  & 4      & 0.94    & 0.94     & 0.94   & 0.87     & 0.85      & 0.81     & 0.63    & 0.79     & 0.70   & 0.83    & 0.80     & 0.72   & 0.83     & 0.80      & 0.72    \\
tuandromd  & 5      & 0.94    & 0.94     & 0.94   & 0.86     & 0.86      & 0.83     & 0.63    & 0.79     & 0.70   & 0.63    & 0.79     & 0.70   & 0.63     & 0.79      & 0.70    \\
tuandromd  & 6      & 0.94    & 0.94     & 0.94   & 0.85     & 0.83      & 0.78     & 0.63    & 0.79     & 0.70   & 0.63    & 0.79     & 0.70   & 0.63     & 0.79      & 0.79    \\ \hline
\end{tabular}
\label{tab:subtab}
\end{sidewaystable}
\subsection{Data Size Imbalance Setting \label{appendix:rawDataDrop}}
\begin{sidewaystable}[]
\centering
\
\caption{\textbf{A.2} Table of precision, recall, and F1 of each model on experimented datasets within 'Data Size Imbalance Setting'. Our CFL able to maintain higher performance. Compared to other models, our CFL gains a significant increase in recall. The precision and F1 scores also improved.}
\scriptsize
\begin{tabular}{|l|l|lll|lll|lll|lll|lll|}
\hline
Model      & Model  & \multicolumn{3}{l|}{Base 1} & \multicolumn{3}{l|}{CFL (Ours)} & \multicolumn{3}{l|}{Base 2} & \multicolumn{3}{l|}{SubTab} & \multicolumn{3}{l|}{SubTab-FL} \\ \hline
dataset    & client & Prec    & Recal    & F1     & Prec     & Recal     & F1       & Prec    & Recal    & F1     & Prec    & Recal    & F1     & Prec     & Recal     & F1      \\ \hline
blog       & 1      & 0.79    & 0.80     & 0.79   & 0.74     & 0.75      & 0.74     & 0.72    & 0.72     & 0.72   & 0.72    & 0.73     & 0.72   & 0.70     & 0.71      & 0.70    \\
blog       & 2      & 0.79    & 0.80     & 0.79   & 0.80     & 0.79      & 0.80     & 0.77    & 0.77     & 0.77   & 0.77    & 0.77     & 0.77   & 0.73     & 0.71      & 0.72    \\
blog       & 3      & 0.79    & 0.80     & 0.79   & 0.81     & 0.80      & 0.80     & 0.80    & 0.79     & 0.79   & 0.79    & 0.78     & 0.78   & 0.77     & 0.75      & 0.75    \\
blog       & 4      & 0.79    & 0.80     & 0.79   & 0.76     & 0.77      & 0.76     & 0.75    & 0.76     & 0.76   & 0.75    & 0.74     & 0.75   & 0.74     & 0.73      & 0.73    \\ \hline
covtype    & 1      & 0.55    & 0.60     & 0.56   & 0.49     & 0.54      & 0.50     & 0.24    & 0.49     & 0.32   & 0.37    & 0.49     & 0.34   & 0.23     & 0.48      & 0.31    \\
covtype    & 2      & 0.55    & 0.60     & 0.56   & 0.56     & 0.56      & 0.50     & 0.48    & 0.49     & 0.36   & 0.51    & 0.49     & 0.34   & 0.48     & 0.49      & 0.34    \\
covtype    & 3      & 0.55    & 0.60     & 0.56   & 0.61     & 0.64      & 0.61     & 0.47    & 0.51     & 0.42   & 0.42    & 0.49     & 0.38   & 0.40     & 0.49      & 0.40    \\ \hline
income     & 1      & 0.82    & 0.83     & 0.81   & 0.77     & 0.79      & 0.75     & 0.56    & 0.75     & 0.64   & 0.76    & 0.78     & 0.73   & 0.56     & 0.75      & 0.64    \\
income     & 2      & 0.82    & 0.83     & 0.81   & 0.74     & 0.76      & 0.68     & 0.66    & 0.75     & 0.65   & 0.72    & 0.76     & 0.68   & 0.56     & 0.75      & 0.64    \\
income     & 3      & 0.82    & 0.83     & 0.81   & 0.73     & 0.76      & 0.72     & 0.56    & 0.75     & 0.64   & 0.72    & 0.75     & 0.67   & 0.56     & 0.75      & 0.64    \\
income     & 4      & 0.82    & 0.83     & 0.81   & 0.80     & 0.80      & 0.76     & 0.78    & 0.77     & 0.71   & 0.81    & 0.79     & 0.75   & 0.76     & 0.77      & 0.69    \\
income     & 5      & 0.82    & 0.83     & 0.81   & 0.80     & 0.81      & 0.79     & 0.78    & 0.78     & 0.73   & 0.80    & 0.81     & 0.79   & 0.78     & 0.77      & 0.69    \\ \hline
sensorless & 1      & 0.17    & 0.16     & 0.15   & 0.51     & 0.49      & 0.48     & 0.27    & 0.10     & 0.05   & 0.10    & 0.14     & 0.10   & 0.10     & 0.09      & 0.02    \\
sensorless & 2      & 0.17    & 0.16     & 0.15   & 0.71     & 0.70      & 0.69     & 0.11    & 0.12     & 0.07   & 0.13    & 0.13     & 0.11   & 0.16     & 0.15      & 0.12    \\
sensorless & 3      & 0.17    & 0.16     & 0.15   & 0.63     & 0.63      & 0.62     & 0.16    & 0.13     & 0.10   & 0.06    & 0.10     & 0.05   & 0.06     & 0.10      & 0.04    \\
sensorless & 4      & 0.17    & 0.16     & 0.15   & 0.46     & 0.47      & 0.46     & 0.14    & 0.12     & 0.08   & 0.08    & 0.11     & 0.07   & 0.05     & 0.10      & 0.04    \\ \hline
syn        & 1      & 0.86    & 0.86     & 0.86   & 0.78     & 0.77      & 0.77     & 0.78    & 0.78     & 0.78   & 0.39    & 0.39     & 0.38   & 0.37     & 0.37      & 0.35    \\
syn        & 2      & 0.86    & 0.86     & 0.86   & 0.78     & 0.77      & 0.77     & 0.77    & 0.77     & 0.77   & 0.42    & 0.42     & 0.42   & 0.39     & 0.39      & 0.38    \\
syn        & 3      & 0.86    & 0.86     & 0.86   & 0.79     & 0.79      & 0.78     & 0.79    & 0.79     & 0.78   & 0.42    & 0.42     & 0.41   & 0.39     & 0.38      & 0.37    \\
syn        & 4      & 0.86    & 0.86     & 0.86   & 0.79     & 0.78      & 0.78     & 0.78    & 0.78     & 0.77   & 0.39    & 0.40     & 0.39   & 0.36     & 0.36      & 0.35    \\ \hline
tuandromd  & 1      & 0.94    & 0.94     & 0.94   & 0.95     & 0.95      & 0.94     & 0.63    & 0.79     & 0.70   & 0.83    & 0.80     & 0.72   & 0.63     & 0.79      & 0.70    \\
tuandromd  & 2      & 0.94    & 0.94     & 0.94   & 0.84     & 0.85      & 0.83     & 0.63    & 0.79     & 0.70   & 0.63    & 0.79     & 0.70   & 0.63     & 0.79      & 0.70    \\
tuandromd  & 3      & 0.94    & 0.94     & 0.94   & 0.90     & 0.90      & 0.89     & 0.63    & 0.79     & 0.70   & 0.63    & 0.79     & 0.70   & 0.63     & 0.79      & 0.70    \\
tuandromd  & 4      & 0.94    & 0.94     & 0.94   & 0.87     & 0.85      & 0.82     & 0.63    & 0.79     & 0.70   & 0.82    & 0.81     & 0.74   & 0.83     & 0.80      & 0.72    \\
tuandromd  & 5      & 0.94    & 0.94     & 0.94   & 0.86     & 0.86      & 0.83     & 0.63    & 0.79     & 0.70   & 0.63    & 0.79     & 0.70   & 0.63     & 0.79      & 0.70    \\
tuandromd  & 6      & 0.94    & 0.94     & 0.94   & 0.84     & 0.83      & 0.78     & 0.63    & 0.79     & 0.70   & 0.66    & 0.79     & 0.70   & 0.63     & 0.79      & 0.70    \\ \hline
\end{tabular}
\label{table:table_client_drop}
\end{sidewaystable}
\subsection{Class Size Imbalance Setting \label{appendix:rawClassDrop}}    

\begin{sidewaystable}[]
\centering
\caption{ \textbf{A.3} Table of precision, recall, and F1 of each model on experimented datasets within 'Class Size Imbalance Setting'. Our CFL able to maintain higher performance. Compared to other models, our CFL gains a significant increase in recall. Precision also improved.}
 \scriptsize
 \begin{tabular}{|l|l|lll|lll|lll|lll|lll|}
\hline
Model      & Model  & \multicolumn{3}{l|}{Base 1} & \multicolumn{3}{l|}{CFL (Ours)} & \multicolumn{3}{l|}{Base 2} & \multicolumn{3}{l|}{SubTab} & \multicolumn{3}{l|}{SubTab-FL} \\ \hline
dataset    & client & Prec    & Recal    & F1     & Prec     & Recal     & F1       & Prec    & Recal    & F1     & Prec    & Recal    & F1     & Prec     & Recal     & F1      \\ \hline
blog       & 1      & 0.79    & 0.80     & 0.79   & 0.74     & 0.76      & 0.74     & 0.72    & 0.72     & 0.72   & 0.73    & 0.74     & 0.67   & 0.71     & 0.73      & 0.64    \\
blog       & 2      & 0.79    & 0.80     & 0.79   & 0.79     & 0.79      & 0.79     & 0.77    & 0.77     & 0.77   & 0.77    & 0.77     & 0.77   & 0.73     & 0.71      & 0.72    \\
blog       & 3      & 0.79    & 0.80     & 0.79   & 0.80     & 0.80      & 0.80     & 0.80    & 0.79     & 0.79   & 0.79    & 0.78     & 0.78   & 0.77     & 0.75      & 0.76    \\
blog       & 4      & 0.79    & 0.80     & 0.79   & 0.76     & 0.77      & 0.77     & 0.75    & 0.76     & 0.76   & 0.75    & 0.74     & 0.75   & 0.74     & 0.73      & 0.73    \\ \hline
covtype    & 1      & 0.55    & 0.60     & 0.56   & 0.49     & 0.54      & 0.49     & 0.24    & 0.49     & 0.32   & 0.40    & 0.50     & 0.36   & 0.24     & 0.49      & 0.32    \\
covtype    & 2      & 0.55    & 0.60     & 0.56   & 0.56     & 0.54      & 0.47     & 0.48    & 0.49     & 0.36   & 0.51    & 0.49     & 0.34   & 0.54     & 0.49      & 0.34    \\
covtype    & 3      & 0.55    & 0.60     & 0.56   & 0.60     & 0.63      & 0.61     & 0.47    & 0.51     & 0.42   & 0.42    & 0.49     & 0.38   & 0.39     & 0.49      & 0.40    \\ \hline
income     & 1      & 0.82    & 0.83     & 0.81   & 0.77     & 0.79      & 0.75     & 0.56    & 0.75     & 0.64   & 0.75    & 0.75     & 0.66   & 0.56     & 0.75      & 0.64    \\
income     & 2      & 0.82    & 0.83     & 0.81   & 0.74     & 0.76      & 0.68     & 0.66    & 0.75     & 0.65   & 0.72    & 0.76     & 0.68   & 0.56     & 0.75      & 0.64    \\
income     & 3      & 0.82    & 0.83     & 0.81   & 0.73     & 0.76      & 0.71     & 0.56    & 0.75     & 0.64   & 0.72    & 0.75     & 0.67   & 0.56     & 0.75      & 0.64    \\
income     & 4      & 0.82    & 0.83     & 0.81   & 0.80     & 0.80      & 0.76     & 0.78    & 0.77     & 0.71   & 0.81    & 0.79     & 0.75   & 0.76     & 0.77      & 0.69    \\
income     & 5      & 0.82    & 0.83     & 0.81   & 0.79     & 0.81      & 0.79     & 0.78    & 0.78     & 0.73   & 0.80    & 0.81     & 0.79   & 0.78     & 0.77      & 0.69    \\ \hline
sensorless & 1      & 0.17    & 0.16     & 0.15   & 0.48     & 0.47      & 0.46     & 0.27    & 0.10     & 0.05   & 0.07    & 0.14     & 0.08   & 0.06     & 0.11      & 0.05    \\
sensorless & 2      & 0.17    & 0.16     & 0.15   & 0.66     & 0.66      & 0.65     & 0.11    & 0.12     & 0.07   & 0.13    & 0.13     & 0.11   & 0.15     & 0.15      & 0.12    \\
sensorless & 3      & 0.17    & 0.16     & 0.15   & 0.60     & 0.61      & 0.60     & 0.16    & 0.13     & 0.10   & 0.06    & 0.10     & 0.05   & 0.06     & 0.10      & 0.04    \\
sensorless & 4      & 0.17    & 0.16     & 0.15   & 0.44     & 0.46      & 0.45     & 0.14    & 0.12     & 0.08   & 0.08    & 0.11     & 0.07   & 0.05     & 0.10      & 0.05    \\ \hline
syn        & 1      & 0.86    & 0.86     & 0.86   & 0.79     & 0.79      & 0.78     & 0.78    & 0.78     & 0.78   & 0.47    & 0.36     & 0.31   & 0.40     & 0.34      & 0.28    \\
syn        & 2      & 0.86    & 0.86     & 0.86   & 0.78     & 0.77      & 0.77     & 0.77    & 0.77     & 0.77   & 0.42    & 0.42     & 0.42   & 0.39     & 0.39      & 0.38    \\
syn        & 3      & 0.86    & 0.86     & 0.86   & 0.80     & 0.79      & 0.79     & 0.79    & 0.79     & 0.78   & 0.42    & 0.42     & 0.41   & 0.39     & 0.38      & 0.37    \\
syn        & 4      & 0.86    & 0.86     & 0.86   & 0.79     & 0.78      & 0.78     & 0.78    & 0.78     & 0.77   & 0.39    & 0.40     & 0.39   & 0.36     & 0.37      & 0.36    \\ \hline
tuandromd  & 1      & 0.94    & 0.94     & 0.94   & 0.84     & 0.84      & 0.80     & 0.63    & 0.79     & 0.70   & 0.80    & 0.80     & 0.72   & 0.84     & 0.79      & 0.70    \\
tuandromd  & 2      & 0.94    & 0.94     & 0.94   & 0.85     & 0.85      & 0.84     & 0.63    & 0.79     & 0.70   & 0.63    & 0.79     & 0.70   & 0.63     & 0.79      & 0.70    \\
tuandromd  & 3      & 0.94    & 0.94     & 0.94   & 0.90     & 0.90      & 0.89     & 0.63    & 0.79     & 0.70   & 0.63    & 0.79     & 0.70   & 0.63     & 0.79      & 0.70    \\
tuandromd  & 4      & 0.94    & 0.94     & 0.94   & 0.87     & 0.85      & 0.81     & 0.63    & 0.79     & 0.70   & 0.82    & 0.81     & 0.74   & 0.83     & 0.80      & 0.72    \\
tuandromd  & 5      & 0.94    & 0.94     & 0.94   & 0.86     & 0.86      & 0.83     & 0.63    & 0.79     & 0.70   & 0.63    & 0.79     & 0.70   & 0.63     & 0.79      & 0.70    \\
tuandromd  & 6      & 0.94    & 0.94     & 0.94   & 0.85     & 0.83      & 0.79     & 0.63    & 0.79     & 0.70   & 0.66    & 0.79     & 0.70   & 0.66     & 0.79      & 0.70    \\ \hline
\end{tabular}
\label{table:class_imbalance}
\end{sidewaystable}
\subsection{Mixed Case Settings \label{appendix:rawMix}}
\begin{sidewaystable}[]
\centering
 \caption{\textbf{A.4} Table of precision, recall, and F1 of each model on experimented datasets within 'Mixed Case Settings'. Our CFL able to maintain higher performance. Compared to other models, our CFL gains a significant increase in recall. The precision also improved.}
\scriptsize
 \begin{tabular}{|l|l|lll|lll|lll|lll|lll|}
\hline
Model      & Model  & \multicolumn{3}{l|}{Base 1} & \multicolumn{3}{l|}{CFL (Ours)} & \multicolumn{3}{l|}{Base 2} & \multicolumn{3}{l|}{SubTab} & \multicolumn{3}{l|}{SubTab-FL} \\ \hline
dataset    & client & Prec    & Recal    & F1     & Prec     & Recal     & F1       & Prec    & Recal    & F1     & Prec    & Recal    & F1     & Prec     & Recal     & F1      \\ \hline
blog       & 1      & 0.79    & 0.80     & 0.79   & 0.74     & 0.75      & 0.74     & 0.72    & 0.72     & 0.72   & 0.72    & 0.74     & 0.72   & 0.70     & 0.73      & 0.70    \\
blog       & 2      & 0.79    & 0.80     & 0.79   & 0.79     & 0.79      & 0.79     & 0.77    & 0.77     & 0.77   & 0.76    & 0.78     & 0.74   & 0.73     & 0.75      & 0.71    \\
blog       & 3      & 0.79    & 0.80     & 0.79   & 0.80     & 0.80      & 0.80     & 0.80    & 0.79     & 0.79   & 0.79    & 0.78     & 0.78   & 0.77     & 0.75      & 0.76    \\
blog       & 4      & 0.79    & 0.80     & 0.79   & 0.76     & 0.77      & 0.76     & 0.75    & 0.76     & 0.76   & 0.75    & 0.74     & 0.75   & 0.74     & 0.73      & 0.73    \\ \hline
covtype    & 1      & 0.55    & 0.60     & 0.56   & 0.49     & 0.54      & 0.50     & 0.24    & 0.49     & 0.32   & 0.38    & 0.48     & 0.34   & 0.23     & 0.48      & 0.31    \\
covtype    & 2      & 0.55    & 0.60     & 0.56   & 0.57     & 0.55      & 0.48     & 0.48    & 0.49     & 0.36   & 0.48    & 0.50     & 0.36   & 0.44     & 0.49      & 0.35    \\
covtype    & 3      & 0.55    & 0.60     & 0.56   & 0.63     & 0.64      & 0.62     & 0.47    & 0.51     & 0.42   & 0.42    & 0.49     & 0.38   & 0.40     & 0.49      & 0.38    \\ \hline
income     & 1      & 0.82    & 0.83     & 0.81   & 0.77     & 0.79      & 0.75     & 0.56    & 0.75     & 0.64   & 0.74    & 0.76     & 0.68   & 0.56     & 0.75      & 0.64    \\
income     & 2      & 0.82    & 0.83     & 0.81   & 0.74     & 0.76      & 0.68     & 0.66    & 0.75     & 0.65   & 0.68    & 0.75     & 0.64   & 0.56     & 0.75      & 0.64    \\
income     & 3      & 0.82    & 0.83     & 0.81   & 0.73     & 0.76      & 0.71     & 0.56    & 0.75     & 0.64   & 0.72    & 0.75     & 0.67   & 0.56     & 0.75      & 0.64    \\
income     & 4      & 0.82    & 0.83     & 0.81   & 0.80     & 0.80      & 0.76     & 0.78    & 0.77     & 0.71   & 0.81    & 0.79     & 0.75   & 0.76     & 0.77      & 0.69    \\
income     & 5      & 0.82    & 0.83     & 0.81   & 0.80     & 0.81      & 0.79     & 0.78    & 0.78     & 0.73   & 0.80    & 0.81     & 0.79   & 0.78     & 0.77      & 0.69    \\ \hline
sensorless & 1      & 0.17    & 0.16     & 0.15   & 0.48     & 0.47      & 0.46     & 0.27    & 0.10     & 0.05   & 0.08    & 0.14     & 0.09   & 0.02     & 0.09      & 0.03    \\
sensorless & 2      & 0.17    & 0.16     & 0.15   & 0.68     & 0.68      & 0.67     & 0.11    & 0.12     & 0.07   & 0.08    & 0.12     & 0.08   & 0.07     & 0.13      & 0.09    \\
sensorless & 3      & 0.17    & 0.16     & 0.15   & 0.63     & 0.63      & 0.62     & 0.16    & 0.13     & 0.10   & 0.06    & 0.10     & 0.05   & 0.06     & 0.10      & 0.04    \\
sensorless & 4      & 0.17    & 0.16     & 0.15   & 0.40     & 0.42      & 0.41     & 0.14    & 0.12     & 0.08   & 0.08    & 0.11     & 0.07   & 0.05     & 0.10      & 0.04    \\ \hline
syn        & 1      & 0.86    & 0.86     & 0.86   & 0.64     & 0.63      & 0.62     & 0.61    & 0.60     & 0.59   & 0.25    & 0.26     & 0.25   & 0.23     & 0.23      & 0.20    \\
syn        & 2      & 0.86    & 0.86     & 0.86   & 0.61     & 0.60      & 0.58     & 0.60    & 0.57     & 0.55   & 0.22    & 0.23     & 0.21   & 0.20     & 0.20      & 0.17    \\
syn        & 3      & 0.86    & 0.86     & 0.86   & 0.62     & 0.62      & 0.61     & 0.59    & 0.58     & 0.56   & 0.26    & 0.26     & 0.25   & 0.22     & 0.22      & 0.21    \\
syn        & 4      & 0.86    & 0.86     & 0.86   & 0.61     & 0.60      & 0.59     & 0.60    & 0.60     & 0.58   & 0.24    & 0.25     & 0.24   & 0.21     & 0.21      & 0.20    \\
syn        & 5      & 0.86    & 0.86     & 0.86   & 0.58     & 0.57      & 0.56     & 0.57    & 0.56     & 0.55   & 0.27    & 0.21     & 0.18   & 0.14     & 0.20      & 0.16    \\
syn        & 6      & 0.86    & 0.86     & 0.86   & 0.62     & 0.62      & 0.61     & 0.60    & 0.58     & 0.57   & 0.27    & 0.22     & 0.19   & 0.15     & 0.20      & 0.17    \\
syn        & 7      & 0.86    & 0.86     & 0.86   & 0.54     & 0.53      & 0.52     & 0.52    & 0.51     & 0.50   & 0.29    & 0.23     & 0.20   & 0.27     & 0.20      & 0.17    \\
syn        & 8      & 0.86    & 0.86     & 0.86   & 0.63     & 0.62      & 0.61     & 0.62    & 0.61     & 0.60   & 0.32    & 0.25     & 0.22   & 0.26     & 0.21      & 0.18    \\
syn        & 9      & 0.86    & 0.86     & 0.86   & 0.62     & 0.62      & 0.60     & 0.61    & 0.57     & 0.55   & 0.27    & 0.27     & 0.26   & 0.24     & 0.24      & 0.23    \\
syn        & 10     & 0.86    & 0.86     & 0.86   & 0.70     & 0.70      & 0.69     & 0.68    & 0.67     & 0.66   & 0.28    & 0.29     & 0.27   & 0.24     & 0.25      & 0.23    \\
syn        & 11     & 0.86    & 0.86     & 0.86   & 0.65     & 0.65      & 0.63     & 0.62    & 0.60     & 0.58   & 0.29    & 0.30     & 0.28   & 0.26     & 0.27      & 0.26    \\
syn        & 12     & 0.86    & 0.86     & 0.86   & 0.62     & 0.61      & 0.60     & 0.59    & 0.59     & 0.58   & 0.30    & 0.31     & 0.29   & 0.27     & 0.27      & 0.26    \\
syn        & 13     & 0.86    & 0.86     & 0.86   & 0.60     & 0.59      & 0.58     & 0.58    & 0.56     & 0.54   & 0.28    & 0.29     & 0.28   & 0.25     & 0.26      & 0.24    \\
syn        & 14     & 0.86    & 0.86     & 0.86   & 0.62     & 0.61      & 0.60     & 0.60    & 0.59     & 0.57   & 0.23    & 0.24     & 0.23   & 0.21     & 0.22      & 0.21    \\
syn        & 15     & 0.86    & 0.86     & 0.86   & 0.63     & 0.64      & 0.63     & 0.61    & 0.60     & 0.59   & 0.26    & 0.27     & 0.26   & 0.23     & 0.24      & 0.23    \\
syn        & 16     & 0.86    & 0.86     & 0.86   & 0.63     & 0.62      & 0.61     & 0.62    & 0.61     & 0.60   & 0.24    & 0.25     & 0.24   & 0.22     & 0.22      & 0.21    \\ \hline
tuandromd  & 1      & 0.94    & 0.94     & 0.94   & 0.89     & 0.89      & 0.87     & 0.63    & 0.79     & 0.70   & 0.63    & 0.79     & 0.70   & 0.84     & 0.80      & 0.71    \\
tuandromd  & 2      & 0.94    & 0.94     & 0.94   & 0.84     & 0.85      & 0.84     & 0.63    & 0.79     & 0.70   & 0.68    & 0.68     & 0.68   & 0.63     & 0.79      & 0.70    \\
tuandromd  & 3      & 0.94    & 0.94     & 0.94   & 0.90     & 0.90      & 0.89     & 0.63    & 0.79     & 0.70   & 0.63    & 0.79     & 0.70   & 0.63     & 0.79      & 0.70    \\
tuandromd  & 4      & 0.94    & 0.94     & 0.94   & 0.86     & 0.85      & 0.81     & 0.63    & 0.79     & 0.70   & 0.82    & 0.81     & 0.74   & 0.83     & 0.80      & 0.73    \\
tuandromd  & 5      & 0.94    & 0.94     & 0.94   & 0.86     & 0.86      & 0.83     & 0.63    & 0.79     & 0.70   & 0.63    & 0.79     & 0.70   & 0.63     & 0.79      & 0.70    \\
tuandromd  & 6      & 0.94    & 0.94     & 0.94   & 0.85     & 0.83      & 0.78     & 0.63    & 0.79     & 0.70   & 0.66    & 0.79     & 0.70   & 0.66     & 0.79      & 0.70    \\ \hline
\end{tabular}
    \label{tab:mixSetup}
\end{sidewaystable}
\end{document}